\documentclass[runningheads]{llncs}
\usepackage[T1]{fontenc}
\usepackage{graphicx}
\usepackage{booktabs}
\usepackage[misc]{ifsym}
\usepackage{amsmath}
\usepackage{amsfonts}
\usepackage{bbm}
\usepackage{bm}
\usepackage{booktabs}
\usepackage{algorithm}
\usepackage{algorithmic}
\usepackage{amsfonts}
\usepackage{graphicx}
\usepackage{wrapfig}
\usepackage{subcaption}
\usepackage{enumitem}
\usepackage{makecell}
\newcommand{\corr}{(\Letter)}
\usepackage{mwe}
\usepackage[hidelinks]{hyperref}

\begin{document}

\title{Informative Semi-Factuals for XAI: \\The Elaborated Explanations that People Prefer}

\titlerunning{Informative Semi-Factuals for XAI}

\author{Saugat Aryal\inst{1,2} \corr \and
Mark T. Keane \inst{1,2}}


\institute{School of Computer Science, University College Dublin, Dublin, Ireland
\and
Insight Research Ireland Centre for Data Analytics, Dublin, Ireland \email{saugat.aryal@ucdconnect.ie} \email{mark.keane@ucd.ie}}

\maketitle              

\begin{abstract}
Recently, in eXplainable AI (XAI), \textit{even if} explanations -- so-called semi-factuals -- 
have emerged as a popular strategy that explains how a predicted outcome \textit{can remain the same} even when certain input-features are altered. For example, in the commonly-used banking app scenario, a semi-factual explanation could inform customers about better options, other alternatives for their successful application, by saying ``\textit{Even if} you asked for double the loan amount, you would still be accepted". Most semi-factuals XAI algorithms focus on finding maximal value-changes to a single key-feature that do \textit{not} alter the outcome (unlike counterfactual explanations that often find minimal value-changes to several features that alter the outcome). However, no current semi-factual method explains \textit{why} these extreme value-changes do not alter outcomes; for example, a more informative semi-factual could tell the customer that it is their good credit score that allows them to borrow double their requested loan. In this work, we advance a new algorithm -- the \textit{informative semi-factuals} (ISF) method -- that generates more elaborated explanations supplementing semi-factuals with information about additional \textit{hidden features} that influence an automated decision. Experimental results on benchmark datasets show that this ISF method computes semi-factuals that are both informative and of high-quality on key metrics. Furthermore, a user study shows that people prefer these elaborated explanations over the simpler semi-factual explanations generated by current methods.

\end{abstract}

\section{Introduction}
In recent years, as Machine Learning models have become more complex and opaque,  eXplainable Artificial Intelligence (XAI) has advanced methods to improve the transparency of these models, making their decisions more interpretable to humans. Indeed, in some jurisdictions explainability is a legal requirement (see e.g. the EU's GDPR \cite{wachter2017counterfactual} and AI Act \cite{EUAIAct2024}).  \textit{``If only"} explanations using counterfactuals have been extensively explored in XAI, as they naturally show human end-users how a model's predictions can change when input-features are altered (often minimally). For example, if I am turned down for a \$70k loan and query the decision, I could be given a counterfactual (e.g., ``If only you had asked for a slightly lower loan of \$68k, you would have been approved'') giving me some recourse on the outcome. \textit{Even if} or semi-factual explanations are related but differ as they show users how a model's predictions can \textit{remain the same} when input-features are altered.  For example, on being refused a \$70k loan, a semi-factual explanation could say ``Even if you asked for a \$10k loan, you would still be refused''; here, even though this explanation seems less helpful, it still tells me something about my negative outcome (namely, that the bank considers me to be a credit risk even for a much smaller loan). As we shall see, semi-factuals are not always unhelpful as they can often inform users about gainful alternatives to outcomes (e.g., telling a farmer that even if they halved their fertilizer use, they would still get the same high crop yield).  Here, we present a novel algorithm for computing semi-factual explanations, one that aims to make them much more informative for people using AI decision-making systems.


\subsection{How Counterfactuals \& Semi-Factuals Differ}

Although semi-factuals are a special case of the counterfactual -- as they tell us about counters to the facts, albeit ones that \textit{do not} change outcomes -- they have different computational constraints and psychological impacts than counterfactuals; semi-factuals appear to serve very different explanatory purposes.  

Typically, good counterfactuals are computed by finding minimal changes to features that alter negative outcomes; they inform us about alternative antecedents that can change a bad outcome (e.g., asking for a lower amount to get a loan).  People spontaneously use counterfactuals to draw attention to enabling conditions leading to a negative event, that they may have overlooked (\cite{byrne2019counterfactuals}; e.g.,``If only it hadn't been raining, the accident wouldn't have happened'').  Furthermore, counterfactuals implicitly assume \textit{ceteris paribus} (i.e.,``other things being equal''); that the altered outcome is solely due to the specified feature changes and is not due to any other hidden/unmentioned feature changes.  This assumption has to hold for the counterfactual to be meaningful and informative.

In contrast, good semi-factuals often hinge on finding \textit{maximal changes} to a feature that lead to the \textit{same outcome} (e.g., asking for a much lower loan and still being refused); they are often designed to rhetorically convince people that a particular feature is not as important as it first seemed. For example, if after asking for a \$70k loan, I am told that I would still be unsuccessful in a \$10k-loan application, the semi-factual is indicating that loan-amount may not be the main factor in the refusal (e.g., maybe credit score is more important?). Accordingly, semi-factuals can weaken people's causal understanding of events (\cite{mccloy2002semifactual}; e.g., changing a farmer's belief that more fertilizer equals higher yields). Furthermore, unlike counterfactuals, semi-factuals do not seem to assume \textit{ceteris paribus}. Rather, implicitly, they seem to invite a consideration of other factors,\textit{ hidden features} that are affecting the outcome.  For instance, if loan-amount is not \textit{the} main factor affecting my refusal, what is?  Am I being rejected because of my age (I am too young) or income level (its too low) or maybe it's my credit rating (its too low). This observation suggests that semi-factual explanations could be made much more informative, if we could identify such hidden features in the decision process and inform people about them.

\begin{figure}[t]
\centering
\includegraphics[width=0.8\textwidth]{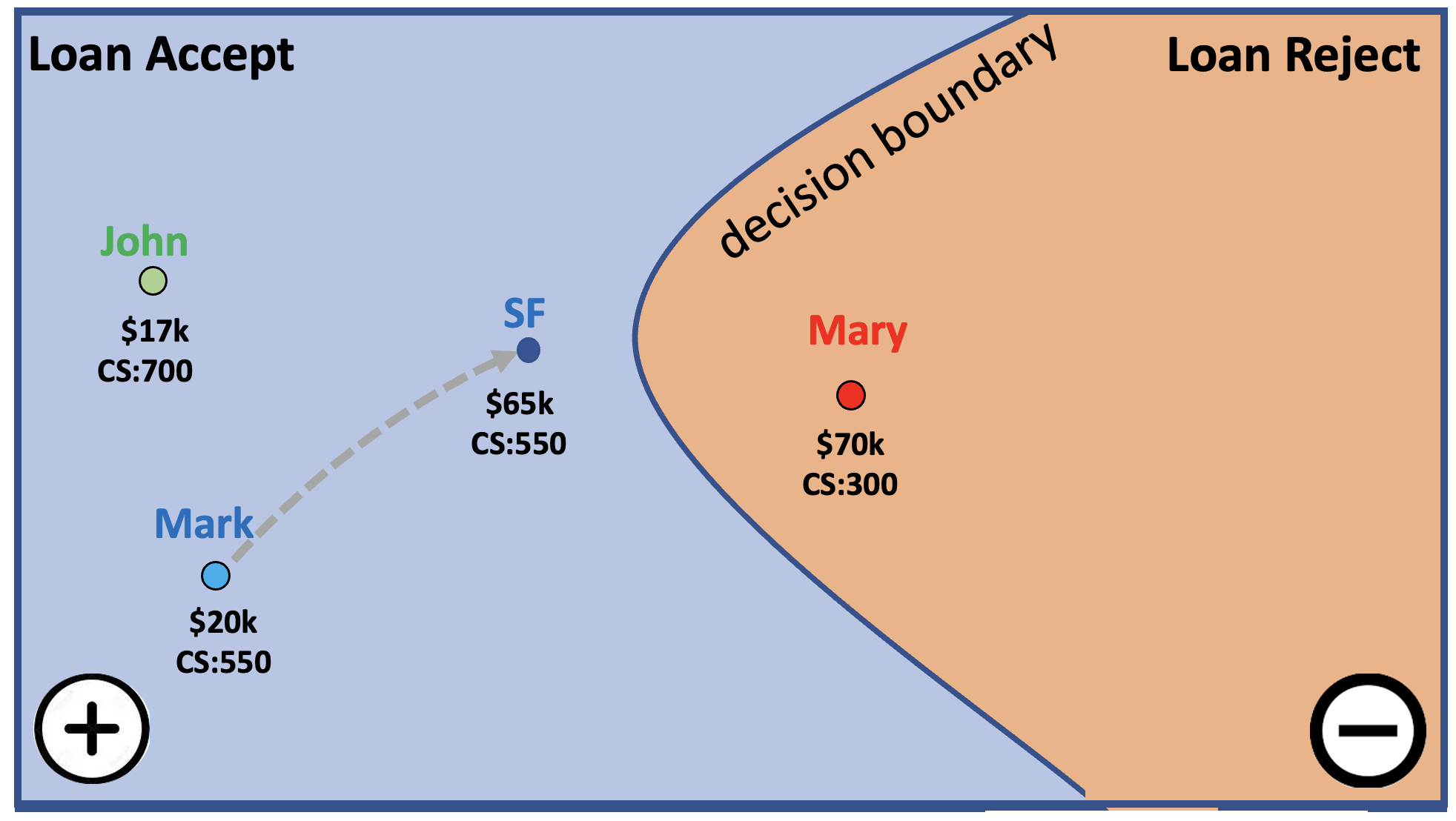}
\caption{The decision space for three loans, with two features shown (i.e., Loan Amount in \$s and Credit Score as CS), for three applicants: John, Mark and Mary. John and Mark have their loans accepted but Mary has been rejected. Mark asks for an explanation about how to get a better deal.  The semi-factual (SF) tells him that with his 550 credit score, he can actually get a \$65k loan. Mark previously thought that if he asked for more, he would end up being rejected like Mary.  The semi-factual shows Mark the limits on his loan aplication given his credit score (n.b., if he asked for \$70k, he would be rejected like Mary). }
\label{fig:sf_bound}
\end{figure}

Figure \ref{fig:sf_bound} shows a loan scenario in which a semi-factual explanation could be made more informative by revealing information about other features. Imagine Mark has two friends -- John and Mary -- who have recently applied for loans to the same bank as Mark. John, who has a decent credit score (of 700), asked for a lowish loan (\$17k) and was successful. Mary was unsuccessful, perhaps because she asks for a higher loan (\$70k) and/or because she has a poor credit score (of 300).
Mark has a low-to-fair credit score (of 550) and when he applies for a modest loan (of \$20k), he is successful. But, he wants to know if he can get a better deal. Could he perhaps borrow more?  Here, the semi-factual explanation could tell him, that even if he asked for a \$65k loan he would be successful, as his credit score is good enough to make him less of a risk than Mary.  At present, no existing semi-factual method considers this additional information and such explanations have not been user-tested. In this paper, we consider both.

\subsection{Computing More Informative Semi-Factuals}

Aryal \& Keane \cite{ijcai2023p732} found that most semi-factual methods compute explanations that have maximal value-changes, typically to a single key feature of a query instance, that leave the outcome unaltered.  For example, their Most Distant Neighbor (MDN) method considers each feature-dimension of the query and tries to find another instance in the dataset that is furthest from it, while still being in the same class. However, none of these methods consider "why" it is possible to radically change a key-feature's value without causing a class-change. So, none of these methods provide users with a \textit{really informative} semi-factual, one that explains the other hidden features influencing the decision outcome (e.g., explaining to Mark that loan amount may not be the most important factor, but that credit score becomes more important). It is the computation of these more informative semi-factuals that is the main focus of this paper.


\begin{figure}[t]
\centering
\includegraphics[width=\columnwidth]{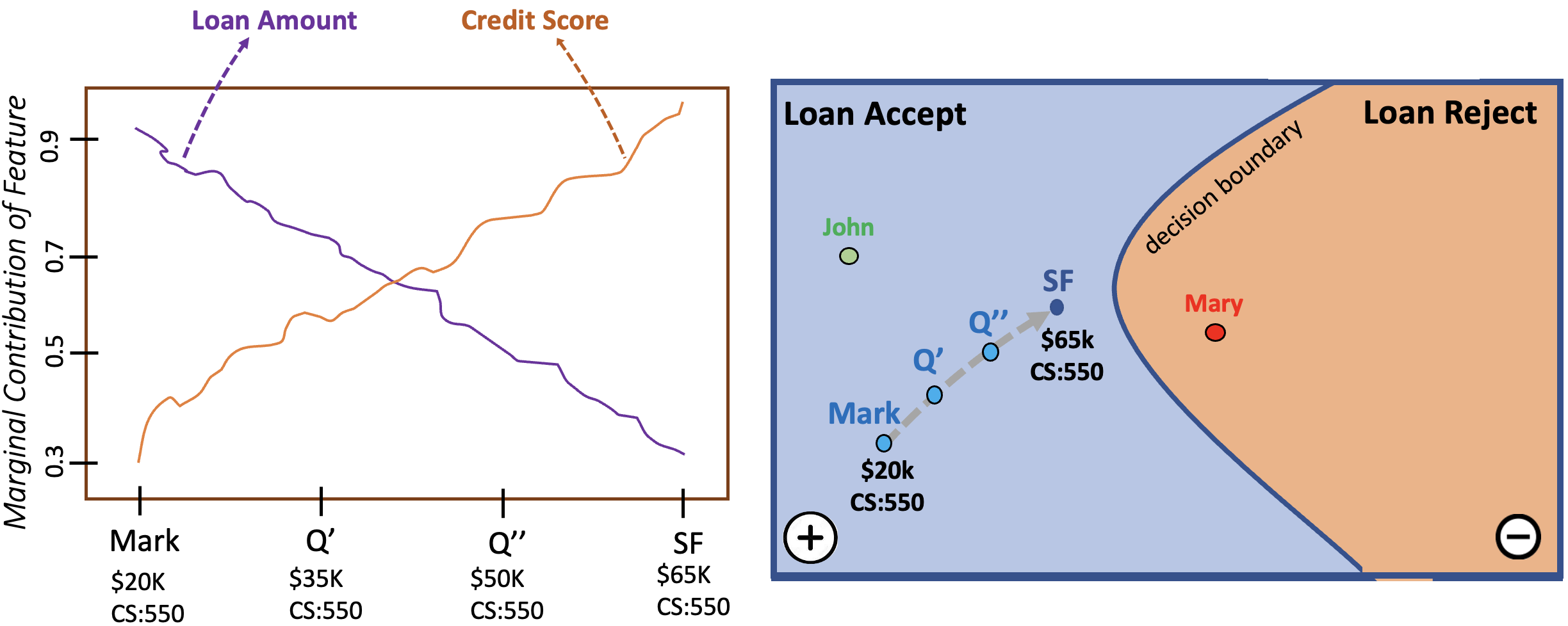}
\caption{The rightmost graph shows a decision space for Mark and his semi-factual explanation (SF), with a path between them based on two perturbation steps (Q' and Q'') in which the key-feature \textit{loan amount} is systematically increased from \$20k to \$65K, without changing \textit{credit-score} (which stays at 550). The leftmost graphic shows the relative changes in the marginal contributions of these two features across these perturbed instances as they remain in the loan-accept class. As \textit{loan amount} increases its marginal contribution to keeping instances in the loan-accept class decreases (see \textit{purple} plot) and even though \textit{credit-score's} value does not change, it's marginal contribution increases (see \textit{orange} plot) revealing this \textit{seesaw pattern} between the two features.}
\label{fig:sf_xp}
\end{figure}

Our hypothesis is that these more informative semi-factuals reflect local changes in the relative importance of different features across different instances in the data distribution.  Imagine a perturbation path between Mark and his semi-factual (SF), created by systematically changing only the value of the loan amount in incremental steps (see right of Figure \ref{fig:sf_xp}).  For any given predictive model, we can assess the marginal contribution of these feature changes for a given instance with respect to it remaining in its current class or moving to another class. Over these perturbation steps, as \textit{loan amount }increases, instances move closer to the decision boundary with the loan-reject class.  But, these changes do not tip it into the loan-reject class because credit-score's influence is simultaneously increasing (see left of Figure \ref{fig:sf_xp}). Even though, inevitably, increasing \textit{loan amount} will eventually flip the class, over these perturbations \textit{credit score} is effectively blocking that class change, as its marginal contribution increases (n.b., even though credit-score's value does not change).  Hence, there is a characteristic \textit{seesaw pattern} in the importance of these two features, as the \textit{loan-amount} key-feature weakens and credit-score hidden-feature strengthens.   
Based on these ideas, we propose a novel explanation algorithm -- the\textit{ Informative Semi-factuals (ISF)} method -- that relies on computing these changing feature contribution patterns in different feature-types as a new constraint in a multi-objective XAI method (see Section 2). We then report some user studies to test whether people actually find these elaborated semi-factuals to be more useful as explanations of automated decisions (see Section 4).




Our intuition was that these patterns of changing importance between key- and hidden-features occur in most good semi-factuals.   To test this intuition we carried out an extensive test on a large sample of the best semi-factuals produced by an ensemble of the main semi-factual methods in the literature (see Appendix A.1).  This dataset of explanations came from comprehensive tests of 8 different semi-factual methods applied to 7 representative tabular datasets (using 5-fold  cross-validation) evaluated against 5 key metrics (i.e., L2-norm distance, plausibility, confusability, robustness, sparsity)  \cite{aryal2024even}.  Based on a large sample of queries from these datasets,  $\sim$10,000 of the best semi-factuals were identified irrespective of the explanation method that produced them (n.b., most were generated-instance explanations rather than existing data-points). This analysis showed that 89\% of the best semi-factuals found show the seesaw pattern of changing marginal contributions between key- and hidden-features (akin to that shown in Fig. \ref{fig:sf_xp}). So, it appears that this property of semi-factuals has always been present in good semi-factuals, and can, therefore, be used to compute more informative semi-factuals for end users. Hence, we proceeded to implemented a new semi-factual explanation method with these new constraints.

\subsection{Outline of Paper \& Contributions}
In the remainder of this paper, we begin by formally defining informative semi-factuals that include new constraints along with their traditional requirements (Section \ref{sec:inform_sf}). Subsequently, we introduce a novel method that computes these informative semi-factuals based on the defined properties (Section 2.2). In Section 3, computational experiments are run to evaluate the performance of this new method in obtaining informative and good semi-factuals by comparison to existing methods. Next, we report two user studies to examine whether people find these elaborated explanations to be more useful (Section 4). Finally, we review the existing works in semi-factual literature (Section 5), before closing with some discussion (Section 6). 

As such, the paper makes three new contributions to the field of semi-factual explanations by (1) defining a new desideratum for semi-factuals, namely the necessity to surface hidden features to provide more informative explanations, (2) proposing a novel method for computing this new requirement, the \textit{Informative Semi-factuals (ISF)} method, along with tests showing it produces the best semi-factuals, (3) reporting two novel user tests showing that people prefer these elaborated semi-factual explanations over ones lacking such elaboration.


    
    
    

\section{Computing Informative Semi-factuals}
Earlier, we advanced the argument for computing more informative semi-factuals, ones that convey a better mental model of the feature contributions leading to various automated decisions.  This argument hinged on the proposal that good semi-factual instances involve a seesaw pattern in which the marginal contribution of key-features weaken (i.e, \textit{key-feature weakening}) as the contribution of hidden-features simultaneously strengthen (i.e, \textit{hidden-feature strengthening}) relative to the presented query instance.  In this section, we propose a new algorithm for computing these more informative semi-factuals  - the Informative Semi-factual (ISF) method -- based on implementing these ideas as new constraints on semi-factual generation.  Accordingly, we need to extend the current desiderata for semi-factuals to formally specify these new constraints (see Definition 2).  Then, armed with this new specification we set about defining a multi-objective optimization method for computing ``better'' semi-factuals, ones that give people more informative explanations (see Section 2.2). 

\subsection{Formalizing Informative Semi-factuals}\label{sec:inform_sf}

Informative Semi-factuals are formalized in the following definitions that are distinguished from previously-proposed semi-factuals by the addition of two new constraints about feature-contribution changes.

\textbf{Preliminaries.} Let $\hat{f}: \mathcal{X}^D \to \mathcal{Y}$ be a prediction model, $\mathcal{X}$ the D-dimensional feature space and $\mathcal{Y}$ a set of desired outcomes. Let $\gamma: [0,1] \to \mathcal{X}$ be a path from $x_q$ to $x_\text{sf}$ in the feature space, with $\gamma(0) = x_q$ and $\gamma(1) = x_\text{sf}$, where $x_q$ is the query and $x_\text{sf}$ is the semi-factual explanation. For a feature $j$, let $\phi_j(t)$ denote its marginal contribution to the prediction $\hat{f}(\gamma(t))$ at point $t$ where $t \in [0,1]$ and  $T_{\phi_j}$ is the trend strength of the marginal contribution measured by:

\begin{equation}
T_{\phi_j}=\int_0^1 \frac{d}{d t} \phi_j(t) dt 
\end{equation}

\begin{definition}[Semi-factual Explanation]
A semi-factual explanation, $x_\text{sf}$ for a query instance $x_q$ is a data point that satisfies the following: (i) $x_\text{sf}$ has the same prediction as $x_q$, (ii) $x_\text{sf}$ differs from $x_q$ on a key-feature dimension, $k$ (ideally 1), (iii) $x_\text{sf}$ is close to $x_q$ along all other features, and (iv) $x_\text{sf}$ is a plausible instance according to the probability distribution $\mathbb{P}_{\mathcal{X}}$.
\end{definition}

\begin{definition}[Informative Semi-factual Explanation]
A given semi-factual explanation, $x_\text{sf}$ (as defined in Definition 1), will be an informative semi-factual explanation if it additionally satisfies two properties: 

\begin{itemize}
    \item Key-feature Weakening: The key-feature's marginal contribution, $\phi_k(t)$ has a decreasing trend when moving from $x_q$ to $x_\text{sf}$ along $\gamma$:  
\begin{equation}
T_{\phi_k} < \epsilon
\end{equation}
where $\epsilon < 0 $ is the trend strength parameter.

    \item  Hidden-feature Strengthening: There exists a non-key hidden-feature, which simultaneously exhibits the maximum increasing trend in its marginal contribution $\phi_j(t)$ when transitioning from $x_q$ to $x_\text{sf}$ along $\gamma$:

\begin{equation}
j^*=\arg \max _{j \in\{1, \ldots, D\} \backslash\{k\}} T_{\phi_j}
\end{equation}
\end{itemize}
    
\end{definition}

\noindent In the next sub-section, we show how these definitions can be implemented in the Informative Semi-factual method. 

\subsection{ISF: The Informative Semi-factual Method}
Taking the above specifications we implement this new Informative Semi-factual (ISF) method using a multi-objective optimization approach where the traditional requirements for a good semi-factual are pitted against the new constraints on patterns of marginal contribution changes. Stated simply, this method tries to balance the generation of semi-factuals with maximal value-change on key-features against the marginal contribution trends of key- and hidden-features. 


The semi-factual generation problem can be translated as following multi-objective minimization task:

\begin{equation}
\begin{aligned}
&\min _{x} F(x):=\min _{x}(\text{-}o_1(x,x_q), o_2(x,x_q))\\
&\text { subject to: }\\
&g_1(\hat{f}(x),\hat{f}(x_q)) \leq 0 \quad \text { and } \quad g_2(x) \leq 0
\end{aligned}
\end{equation}

\noindent Here, the first objective $o_1$ ensures maximum distance between $x$ and $x_q$ along key-feature dimension, $k$: 

\begin{equation}
o_1(x, x_q) = |x^k - x_q^{k}|
\end{equation}

\noindent Since, the goal is to maximize this distance, we minimize -$o_1$. The second objective $o_2$ ensures minimizing the cumulative distance across all features except the key feature, encouraging similarity on non-key features:

\begin{equation}
o_2(x, x_q) = \sum_{i \ne k}|x^i - x^i_q| \quad \forall i \in D
\end{equation}

\noindent This $g_1$ constraint ensures that $x$ and $x_q$ have the same class:

\begin{equation}
g_1(\hat{f}(x),\hat{f}(x_q))=\mathbb{I}(\hat{f}(x) \neq \hat{f}(x_q))=0
\end{equation}

The second constraint, $g_2$ ensures that $x$ remains within a plausible region of the data distribution. To do so, we introduce a constraint such that the Probability Density Function (PDF) of $x$ lies within a dynamic threshold range. The range is based on the mean of the probability distribution of the observed data adjusted by an adaptive factor, which scales with the variability of the data:

\begin{equation}
g_2(x)=\mathbb{I}[(\mu - \Delta\delta) \leq \log \mathbb{P}(x) \leq (\mu + \Delta\delta)]
\end{equation}

\noindent where $\mu$ is the mean log-pdf of the data distribution, $\mathbb{P}_{\mathcal{X}}$ and $\Delta\delta$ is an adaptive threshold range, defined as:

\begin{equation}
    \Delta\delta = \theta * \sigma
\end{equation} 

\noindent where $\sigma$ is the standard-deviation of the log-pdf of the data distribution and $\theta$ is a scaling parameter which controls the strictness of the constraint. 
 
This optimization method produces a set of diverse and equally valid semi-factuals. Next, we analyze the marginal contribution of the features in these explanations to obtain informative semi-factuals. To do so, for each of the explanation-solutions, we linearly interpolate from $x_q$ to $x_\text{sf}$ as: 

\begin{equation}
x(t) = (1-t) x_q+t x_\text{sf}
\end{equation}

\noindent where  $t \in [0,1]$ and $x(t)$ is an instance during interpolation at point $t$ with $t = 0$ at $x_q$ and $t = 1$ at $x_\text{sf}$. During each step of interpolation, the marginal contribution of all the features (both key and non-key) on the outcome are noted. 
Specifically, their pure main effects, $\phi_j(t)$, are determined to obtain a good estimation of how feature-influences are changing independently during interpolation. Trends in marginal contributions are identified using the Kendall's tau of Mann-Kendall test \cite{mann1945nonparametric}. Finally, we select those instances in which the key-feature's contribution has the lowest decreasing trend as the best candidate semi-factual explanation. Concurrently, for this selected explanation, we identify the non-key feature with the highest increasing trend as the hidden-feature component, to form the most \textit{informative} semi-factual explanation.

For a given query, we run the method by treating each of its features as a key-feature to obtain multiple informative semi-factuals across each feature-dimension before selecting the best of the best as the final informative explanation that has the overall lowest key-feature-weakening trend.

\section{Testing ISF Algorithm}

Two computational experiments compared ISF's performance on representative datasets to the performance of an ensemble of leading semi-factual methods.  The first experiment determined whether ISF's semi-factual generation process correctly produced explanations with the requisite properties (i.e., the key-feature weakening and hidden-feature strengthening pattern) relative to ensemble methods that were not specifically designed to compute these properties. The second experiment evaluated the semi-factuals generated by ISF by comparison to the best semi-factuals generated by the ensemble methods to determine their relative goodness on the traditional metrics. 
In both experiments a leave-one-out cross-validation was performed in which each query instance was used to generate candidate semi-factuals for every  dataset (for source code and data see \textit{\href{https://github.com/anonymous-sf/informative-semifactuals-ecml2026}{here}}).  Taken together these experiments determine whether (i) ISF really does what it claims to do in generating informative explanations, and (ii) ISF produces quality semi-factual explanations that are competitive with the SOTA.

\textbf{Setup: Datasets \& Benchmarks.}  Both experiments used five benchmark, publically-available tabular datasets, all of which were binary-classed: Adult Income, Blood Alcohol, PIMA Diabetes, German Credit, and HELOC. Following \cite{van2021interpretable}, all categorical features in these datasets were encoded into a  numerical space using distance metrics. ISF's performance was benchmarked against the best explanations produced by an ensemble of leading semi-factual methods: the  CBR-based Local-Region method \cite{nugent2009gaining}, Knowledge-Light Explanation-Oriented Retrieval (KLEOR)\cite{cummins2006kleor}, Diverse Semi-factual Explanations of Reject (DSER) \cite{artelt2022even}, PIECE \cite{kenny2021generating}, C2C-VAE \cite{zhao2022generating}, MDN\cite{ijcai2023p732}, S-GEN\cite{kenny2023sf}, and DiCE \cite{mothilal2020explaining}.  PIECE and C2C-VAE had to be modified to work with tabular data. For DiCE, the desired class of the explanation was set to be the same as that of the query, to allow it to generate semi-factuals not counterfactuals. Appendices A.1, A.2 have full description of the models, the modifications made to them and parameters used.

\textbf{ISF Implementation.}  
ISF implements the semi-factual generation as a multi-objective minimization task using Equations (4)-(10). We used Gaussian copula \cite{nelsen2006introduction} 
to model the joint distribution of features and obtain $\mathbb{P}_{\mathcal{X}}$.
In Eq.(9) the threshold was set as $\theta=1.5$.  The \textit{Nondominated Sorting Genetic Algorithm II} (NSGA-II) \cite{deb2002fast} was used to solve the constrained multi-objective semi-factual problem. The algorithm efficiently approximates the Pareto front to produce a set of diverse and equally-valid semi-factual solutions.  We used an initial population size of 50 for 100 generations (see Appendix A.3 for full details on parameters).
A Random Forest Classifier model was used to determine class-membership (i.e., as in the $g_1$ constraint). The pure marginal contribution of features was computed using the diagonal entries of the SHAP interaction matrix \cite{lundberg2017unified} as it isolates the main effects. Specifically, the TreeSHAP variant \cite{lundberg2020local2global} was used to  work with the Random Forest model. To compute the changes in marginal contributions, we created 10 intermediate instances interpolating from the query to the generated semi-factual. The trend in feature contributions occurring over these perturbed instances was analyzed using the Mann-Kendall trend test.
The resulting Kendall's tau ($\tau$) coefficient was used to measure trend strength, where $\tau \in [-1,1]$ indicates the direction and strength of the trend. We set $\epsilon=-0.3$ in Eq.(2) as the trend strength for key-feature weakening.


\subsection{Experiment 1: Informative Semi-factuals?}

\begin{figure}[t]
\centering
\includegraphics[width=0.7\linewidth]{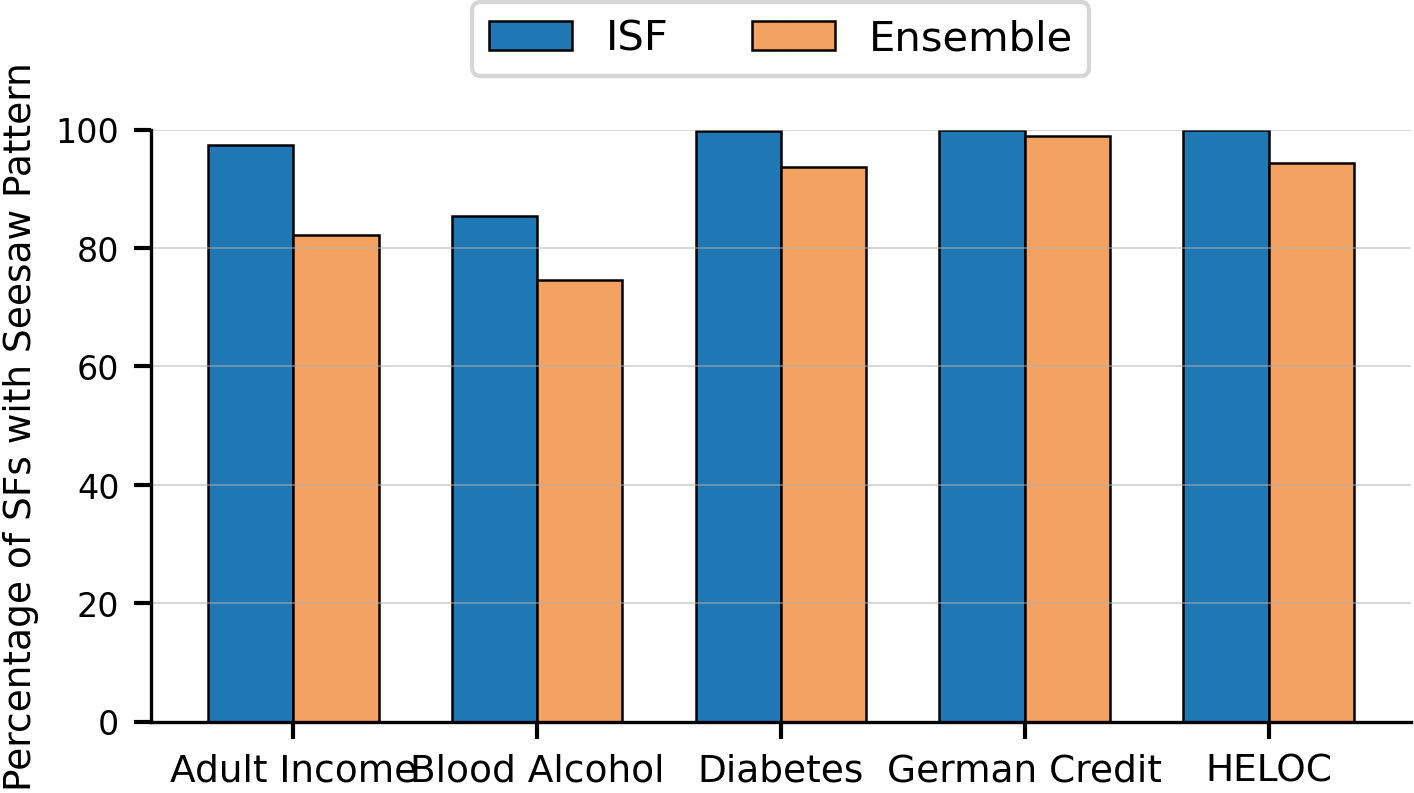}
\caption{From Expt.1, the percentage of semi-factuals, for five datasets, generated by the ISF and Ensemble-methods (N=38,233 in total), that manifested the seesaw pattern in key-feature versus hidden-feature contributions.}
\label{fig:expt1}
\end{figure}

\begin{figure}[t]
    \centering
    
    \begin{subfigure}{0.5\textwidth}
        \centering
        \includegraphics[width=0.6\linewidth]{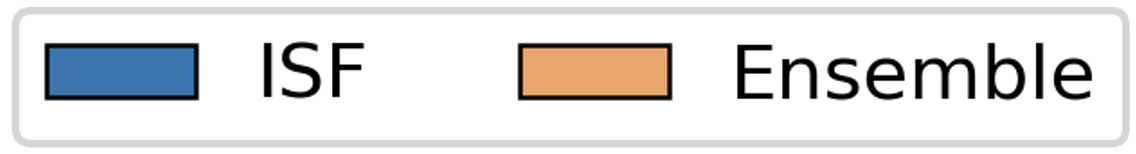} 
    \end{subfigure}
    
    \begin{subfigure}{0.48\textwidth}
    \centering
        \includegraphics[width=\linewidth]{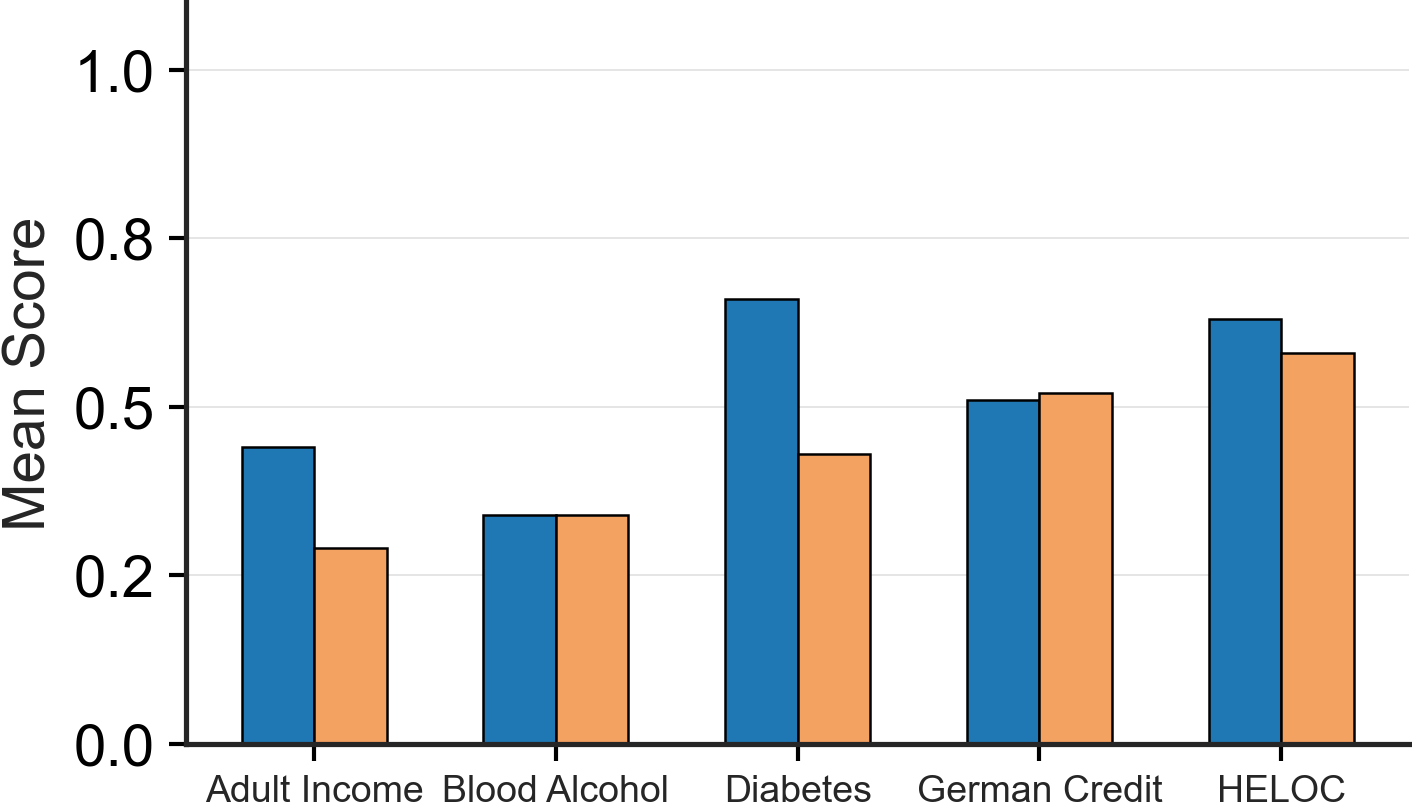}
        \caption{Distance}
        \label{fig:distance}
    \end{subfigure}
    \hfill
    \begin{subfigure}{0.48\textwidth}
        \includegraphics[width=\linewidth]{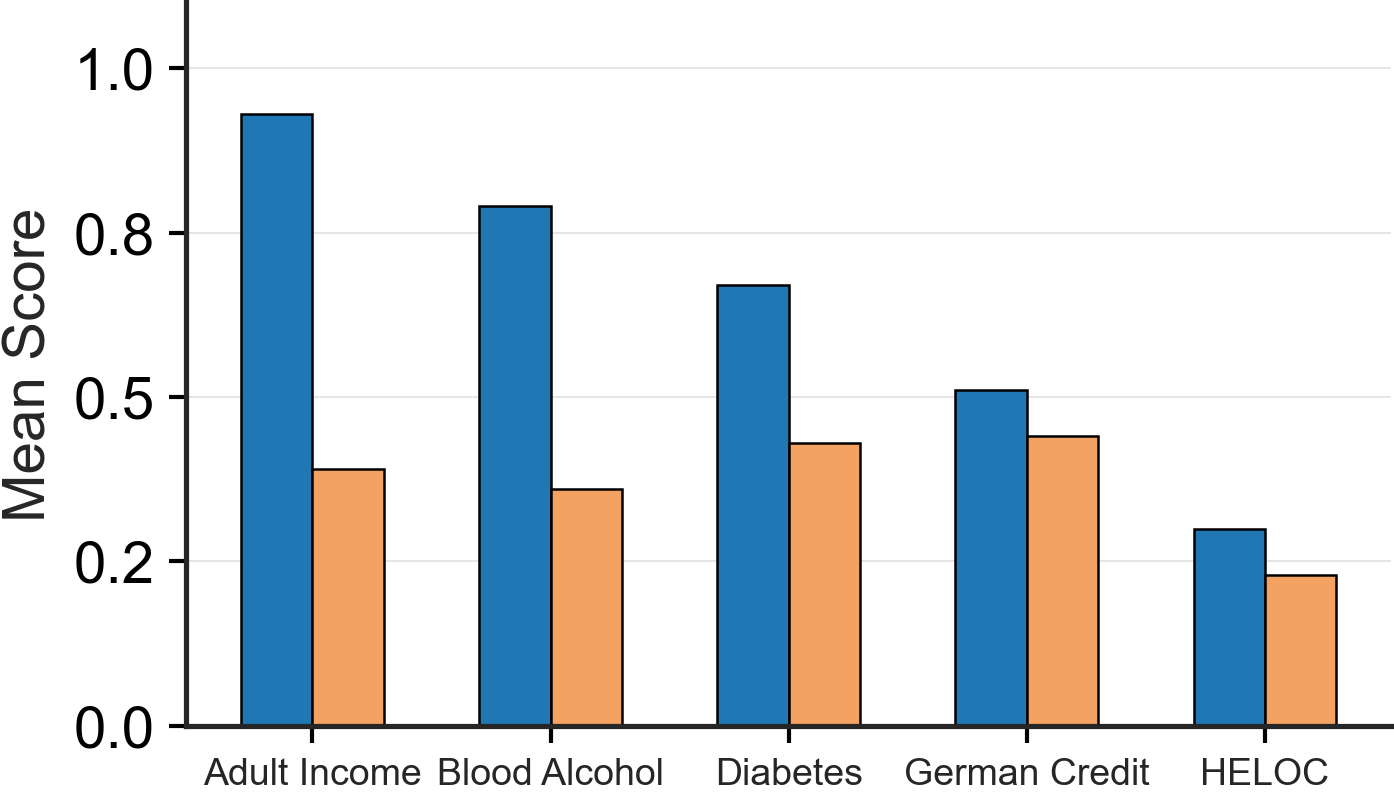}
        \caption{Sparsity}
        \label{fig:sparsity}
    \end{subfigure}
    
    \begin{subfigure}{0.48\textwidth}
        \includegraphics[width=\linewidth]{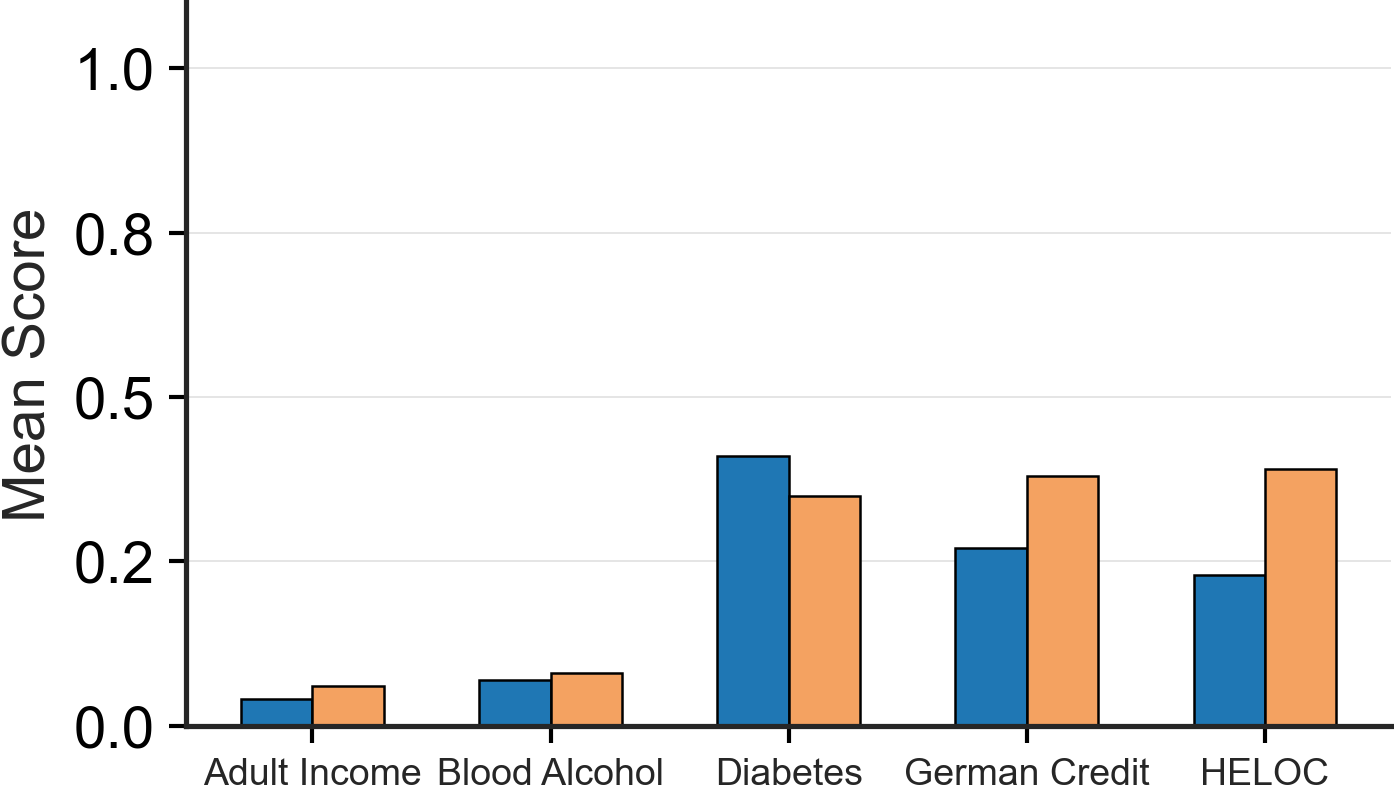}
        \caption{Plausibility}
        \label{fig:plausibility}
    \end{subfigure}
    \hfill
    \begin{subfigure}{0.45\textwidth}
        \includegraphics[width=\linewidth]{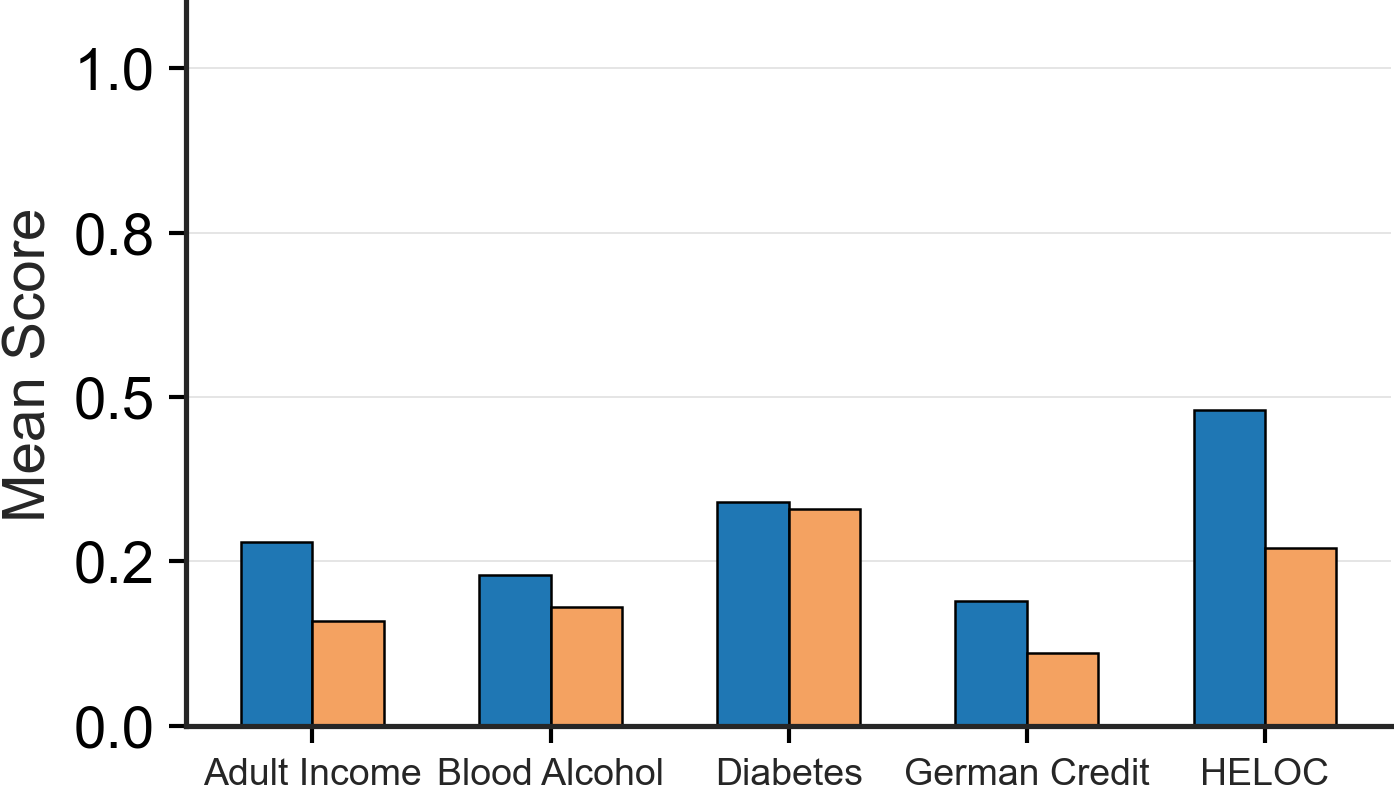}
        \caption{Trustworthiness}
        \label{fig:trust}
    \end{subfigure}

    \caption{From Expt.2, the mean goodness scores for semi-factuals generated by the ISF and Ensemble-methods for the five datasets (N=38,233 in total), in each of the four evaluation metrics (a-d).}
    \label{fig:expt2}
\end{figure}

Experiment 1 was designed to determine whether ISF really does what it claims to do in generating informative explanations; namely, does it identify semi-factuals with the seesaw pattern showing a weakening of the key-feature alongside a strengthening hidden-feature.   If these properties are properly identified then ISF will be able to provide informative explanations (e.g., as in ``Mark, you can get a \$65k loan, as you have a fair credit score''). 

\textbf{Procedure.}  Across 5 datasets, a total of 38,223 queries were tested, recording the semi-factuals generated by ISF and the ensemble. For a given query, the best semi-factual from the ensemble models was selected to compare to that generated by ISF.  Note, although none of the ensemble models explicitly compute patterns of marginal contributions, earlier tests showed that they tend to generate semi-factuals with these properties (though perhaps not optimally so). The measure used was the percentage of semi-factuals generated by ISF/ensemble that showed the seesaw pattern in the changing marginal contributions.

\textbf{Results. } Figure \ref{fig:expt1} shows the percentage of semi-factuals generated by ISF and the ensemble that showed the requisite seesaw pattern needed to compute informative semi-factual explanations.  It shows that across all datasets ISF is able to produce more informative semi-facutals than the ensemble of leading methods, at very high levels of almost 100\% in 4 out of 5 datasets.

\subsection{Experiment 2: Informative Semi-factuals Are Also Good?}

In Experiment 1, we saw that ISF regularly generates semi-factuals with the requisite seesaw pattern needed for more informative explanations.  Using the same setup and procedure, Experiment 2 aimed to determine whether ISF's semi-factuals were also the best semi-factuals relative to those produced by the ensemble of methods. So, it evaluated the quality of semi-factuals generated by ISF and the ensemble using standard metrics for assessing semi-factual goodness \cite{ijcai2023p732,aryal2024even}.  Four commonly-used evaluation metrics were applied to all semi-factuals: 

\begin{itemize}
    \item \textit{Distance:} measures the $L_2$-norm distance between a query and its semi-factual where higher value is preferred.
    \item \textit{Sparsity:} measures feature differences between the query and semi-factual as a ratio of the desired- (set to 1) to observed-difference, where a higher value is better.
    \item \textit{Plausibility:} is measured as the distance between a semi-factual and the nearest training datapoint, where smaller value is better. 
    \item \textit{Trustworthiness:} measures the confidence for a semi-factual for being in the query class compared to the counterfactual class, measured as the ratio of their class distances (normalized), where higher score is better.
\end{itemize}

\textbf{Results. }Figure \ref{fig:expt2} shows that ISF produces good semi-factuals, that are consistently better than those produced by the ensemble of leading methods on key evaluation metrics. It consistently outperforms the ensemble on Sparsity and Trustworthiness for all datasets. ISF's semi-factuals are sparse as they focus on changing only one key-features and are trusted to be classified as being within the query class. These higher trust scores probably occur because of support from hidden features that keep the semi-factual in the query class. ISF also does better on Distance and Plausibility measures across most datasets. Overall, it is clear that ISF produces both informative and high-quality semi-factual explanations, ones that are generally better than the SOTA in the field.


\begin{figure}[ht]
\centering
\begin{subfigure}{0.48\textwidth}
\centering
        \includegraphics[width=\linewidth]{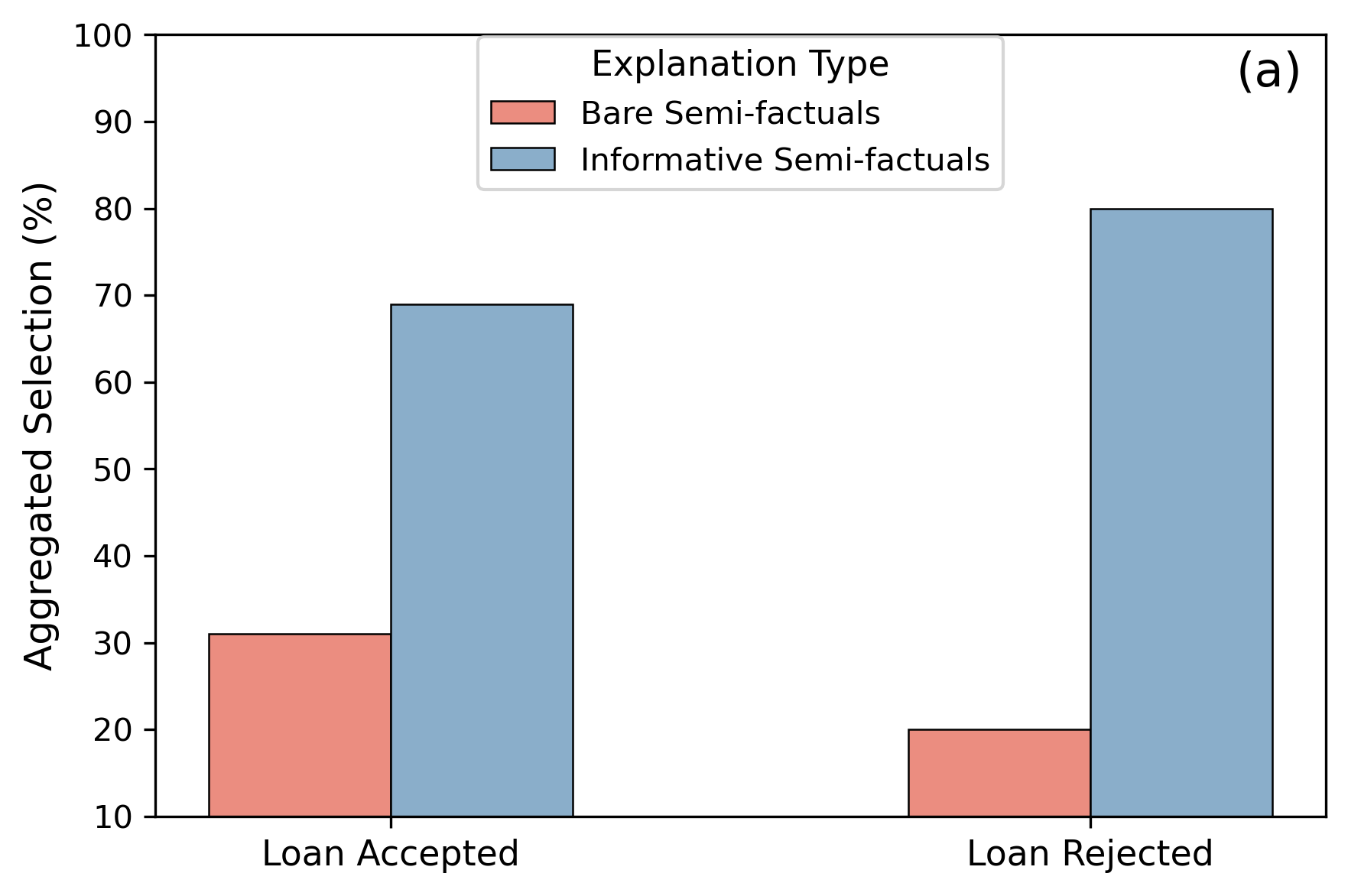}
        \label{fig:test1}
    \end{subfigure}
    \hfill
    \begin{subfigure}{0.48\textwidth}
    \centering
        \includegraphics[width=\linewidth]{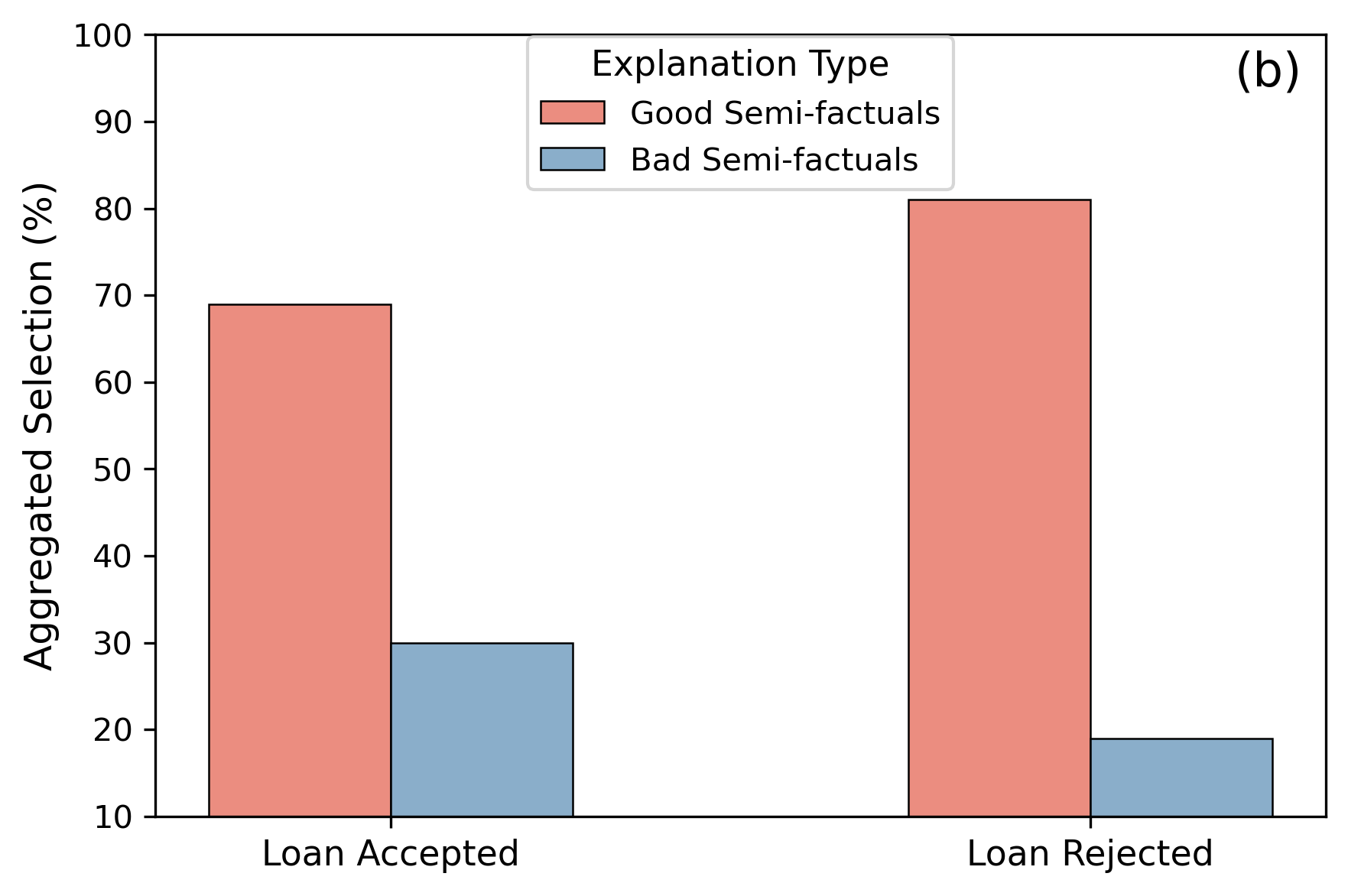}
        \label{fig:test2}
    \end{subfigure}
    \caption{The percentage of people choosing the (a) Bare Semi-factual or Informative Semi-Factual in User Study 1, and (b) the Good or Bad Semi-factuals in User Study 2, for scenarios with the loan-accepted or loan-rejected outcomes}
    \label{fig:user}
\end{figure}

\section{Testing People's Preferences for Explanations}
The computational experiments showed that ISF can generate very good, informative semi-factuals, where we define an ``informative'' semi-factual to be one showing the key-feature and its supporting hidden feature (e.g., ``Even if you asked for \$65k you would be successful, as you have a fair credit score'').   However, user studies are also required to determine whether people think ISF's explanations are acceptable and useful.   Two psychological experiments were carried out to assess this question, in which people assessed the utility of semi-factuals in a forced-choice task; that is, they were shown several loan-scenarios each of which had two semi-factual explanations and asked to select the one they thought was the most useful. Study 1 gave people scenarios with and without the hidden-feature added to the explanation. Study 2 gave people scenarios with semi-factual explanations lacking hidden-features, but which were deemed to be good or bad by the virtue of key-feature's decreasing trend. 

\subsection{User Study 1: Do People Prefer Informative Explanations?}
To our knowledge, Kenny \& Huang's \cite{kenny2023sf} study examining people's preferences for counterfactuals and semi-factuals is the only XAI user test in this area; so, we adapted their design to test the ISF method.  User Study 1 tested whether people prefer bare semi-factual explanations or the more elaborated ones generated by ISF (i.e., mentioning the key feature and the hidden feature). Participants were presented with loan-scenarios for different individuals, each of which had five features (Duration, Credit Amount, Age, Installment Percent, Existing Loans) with loan-accepted/loan-rejected outcomes. For each scenario, they were shown two explanations (counterbalanced in order of presentation across items), one with the semi-factual on its own (Bare-SF; e.g., ``Even if you asked for \$65k you would be successful'') and one showing the semi-factual with the added hidden feature (Informative-SF; e.g., ``Even if you asked for \$65k you would be successful, as you have a fair credit score'').  They were then asked to select the explanation they thought was the most useful as feedback to the customer.

\textbf{Method: Participants, Design, Materials \& Procedure.}
Participants ($N = 15$, based on a power analysis for the single-group design) were recruited from Prolific.com and were pre-screened to be native English speakers from Ireland/UK/USA/Australia/Canada/New Zealand who had not participated in previous related studies. They were paid \textit{\pounds}14/hr for their participation. The study employed a 2 (Explanation Type: Bare-SF v Informative-SF) X 2 (Loan Outcome: Loan Accepted v Loan Rejected) within-subject design. Each participant was presented with 32 different scenarios, half with loan-accepted and half with loan-rejected outcomes. Each scenario was presented with the two versions of the explanation, with and without the hidden feature. For each scenario the participants had to select the explanation they thought was the most useful. For analysis, the count for either choice for both loan outcome scenarios were counted, to be expressed as a percentage.

\textbf{Results.  }Figure 5(a) shows that people find the informative semi-factual explanations with the supporting hidden-feature to more useful than those without that feature (69\% v. 31\% in loan accept and 80\% v. 20\% in loan reject scenarios). Binomial tests further confirmed that these preference differences is statistically significant with $p < .0001$ for both loan outcome scenarios.

\subsection{User Study 2: Preferring Good or Bad Explanations?}
Most semi-factual algorithms can be tweaked to produce good or bad explanations, but we know of no user study that has explicitly tested whether people consider such explanations to be better or worse than one another. User Study 2 tested whether semi-factual explanations deemed to be good/bad by ISF were also deemed to be good/bad by people.  Using the same loan materials, ISF's parameters were modified to produce semi-factuals with high or low scores on key evaluation metrics.   These semi-factuals were then presented as in User Study 1, with people being asked to select the one they thought was the most useful.  To make this a simple test, the explanations were presented as unelaborated, bare semi-factuals (even though, according to ISF, they had different hidden-feature patterns).  So, all of these explanation-pairs used the same key-feature and were classed as good or bad by having distant/close values for this key-feature and by having strong/weak see-saw patterns in their hidden features, respectively.



\textbf{Method: Participants, Design, Materials \& Procedure.}
A new sample of participants ($N = 15$) were recruited using the same criteria and compensation as User Study 1. The experimental design and procedure remained the same, with the only difference being the Explanation Type. 
The study employed a 2 (Explanation Type: Good-SF v Bad-SF) X 2 (Loan Outcome: Load Accepted v Loan Rejected) within-subject design. Each participant was presented with 32 different scenarios, half with loan-accepted and half with loan-rejected outcomes.   Furthermore, each scenario had ``good'' and ``bad'' semi-factuals from which the participants had to select the one they found the most useful. 
The good and bad semi-factuals were obtained from ISF by setting the trend parameter within $-1 \le \epsilon \le -0.8$ and $-0.6 \le \epsilon \le -0.3$ in Eq. (2) respectively.



\textbf{Results.  }Figure 5(b) shows that people find ISF's good semi-factuals to be significantly more useful than its bad semi-factuals (69\% v. 31\% in loan accept and 81\% v. 19\% in loan reject), discriminating their relative goodness as explanations, even when their underlying hidden-features are not provided. 
So, even when people encounter semi-factual explanations without hidden-feature elaborations, at some level, they may ``know'' that there is another factor at work (and have some sense of the strength of that other factor).  

\section{Related Work}
%
While there are no previous studies which considers computing informative semi-factuals, we review some of the key works in semi-factual literature from computational and psychological standpoint  (see \cite{ijcai2023p732} for comprehensive review). 

\textbf{Computational XAI Research. }
Research on semi-factual explanations in XAI originated in early 2000s in the field of Case-Based Reasoning (CBR)  \cite{nugent2005best,nugent2009gaining,cummins2006kleor}. This work characterized semi-factuals as \textit{"a fortiori"} arguments to provide better and more convincing explanations. They used utility functions \cite{doyle2004explanation}, similarity methods \cite{cummins2006kleor}, and local-region based proxy models \cite{nugent2009gaining} to compute such explanations. Kenny \& Keane \cite{kenny2021generating} revisited semi-factuals more systematically extending to generative AI, using GANs based on the exploitation of "exceptional features". This work was followed by Zhao et al. \cite{zhao2022generating}, who used class-to-class variational autoencoders (VAEs) to compute semi-factuals efficiently and Vats et al. \cite{vats2022changes} who used the latent space of these generative models to obtain semi-factuals to explain the classification of medical images such as ulcers.   Aryal \& Keane \cite{ijcai2023p732} surveyed the semi-factual literature to define the "requirements" for cognitively and computationally good semi-factuals. They proposed a novel benchmark method, Most Distant Neighbors (MDNs), which used a scoring function to select the most distant same-class instance as the semi-factual. In other work, Aryal \& Keane \cite{aryal2024even} demonstrated that semi-factuals are conceptually and computationally distinct from counterfactuals, possessing their own unique dynamics rather than being mere "by-products". Kenny \& Huang \cite{kenny2023sf} advanced the research further by introducing "gain" as a new constraint for semi-factual explanations. They argued that semi-factuals can provide better recourses for positive outcomes whereas counterfactuals are more useful for negative outcomes, which they validated through user studies.

\textbf{Other Computational Research. } Semi-factuals have also been used for model auditing; specifically for identifying spurious patterns and bias \cite{lu2022rationale} or for providing unified explanatory frameworks via Gaussian mixture models \cite{xie2023joint}. Similarly, Artelt \& Hammer\cite{artelt2022even} utilized semi-factuals to justify "reject" decisions made by a model. Dandl et al. \cite{dandl2023interpretable} proposed Interpretable Region Descriptors (IRDs) for semi-factuals which maps the stable region around the query through "hyperboxes". More recently, the scope of semi-factual explanations have expanded into dynamic environments. Gajcin et al. \cite{gajcin2024semifactual} successfully adapted semi-factual desiderata to Reinforcement Learning (RL) agents using genetic algorithms to identify stable states. Jiang et al. \cite{jiang2024interpreting} generated semi-factual explanations for Reward Models (RMs) to analyze their local and global behavior.

\textbf{Psychological Research. }The research on the cognitive effects of semi-factuals were originally carried out in the field of psychology. The seminal work of McCloy \& Byrne \cite{mccloy2002semifactual} found that semi-factual can weaken the causal relation between an input and outcome, convincing people that the outcome would have occurred regardless. Similarly, they also showed that semi-factual thoughts decrease the emotion of regret. On the other hand, Green \cite{green2008persuasion} showed that a semi-factual supports dissuasion convincing people to \textit{not} take further action.

\section{Discussion}
Semi-factual explanations employ \textit{Even if} reasoning to explain how changes to certain input-features do not change the outcome. In XAI, all the existing works on semi-factuals have focused on computing the farthest instances from the query, within the same class, using various different techniques. However, to the best of our knowledge, none of these models examine "why" it is possible for a semi-factual to have such maximal value-changes and yet remain in the same class. This work advances the research on semi-factual explanations by introducing the notion of \textit{informative semi-factuals}. These explanations complement a standard semi-factual with \textit{hidden features} that further explain how the semi-factual can occur in the data distribution of a class. To compute these informative semi-factuals, we proposed a novel -- the Informative Semi-factual (ISF) method -- which performs better than the SOTA methods in computational experiments. We also conducted  user studies which shows that people find these elaborated semi-factual explanations better and more useful than the simple standard ones. A notable limitation of our work is its computational efficiency when there are large number of features to consider in the multi-objective setting. However, this may be addressed with feature-selection techniques that identify the most relevant or locally-sparse features. As part of future work, it would also be interesting to explore how this concept can be formalized in other domains such as images, and time series which may require specific considerations. 

\begin{credits}
\subsubsection{\ackname} This work has emanated from research conducted with the financial support of Taighde Éireann – Research Ireland under Grant number: 12/RC/2289\_P2.

\subsubsection{\discintname}
The authors have no competing interests to declare that are relevant to the content of this article.
\end{credits}
%
%
%
\bibliographystyle{splncs04}
\bibliography{references}
%

\begin{appendix}
\section{Appendix}

\subsection{Test for Seesaw Pattern in Best Semi-factuals}
In this test, we computed the presence of seesaw pattern of marginal feature contributions (key- and hidden-features) in best semi-factuals obtained from the ensemble of standard semi-factual methods. 

\textbf{Methods.} The following 8 SOTA semi-factual methods were considered for the ensemble. 

\begin{itemize}[itemsep=2ex]
    \item \textbf{Local-Region Model} Nugent et al. \cite{nugent2009gaining} proposed the Local-Region model, which analyses the local region around the query, using a surrogate model (akin to way  LIME works), to select the nearest neighbor in the query class with the most marginal probability, as the semi-factual instance; as follows:

\begin{equation}
Local\text-Region(q, C)=\arg \min_{x \epsilon C} LR(x)
\end{equation}

where, \emph{C} is the set of candidate neighbors and \emph{LR()} is the local logistic regression model providing the probability score.

    \item \textbf{Knowledge-Light Explanation-Oriented Retrieval (KLEOR)} Cummins \& Bridge \cite{cummins2006kleor} identified query-class instances which lie between the query and its Nearest Unlike Neighbors (NUNs) (aka its counterfactuals) using distance measures and selected the best semi-factual as the one which is closest to the NUN and furthest from the query. They proposed three different variants out of which Attr-Sim was the most advanced. It enforces similarity across all of the features between the query and the candidate instance. 
    
\begin{equation}
\begin{split}
{Attr\text-Sim}(q, nun, G) &= \arg \max_{x \epsilon G} Sim(x,nun) \\
&\qquad {}+ \max_{a \epsilon F} count[Sim(q_a, x_a) > Sim(q_a, nun_a)]
\end{split}
\end{equation}

where \textit{q} is the query, \textit{x} is a candidate instance, $G$ is the set of query-class instances, \textit{nun} is the NUN, \textit{Sim} is Euclidean Distance and \textit{F} is the feature set. 

    \item \textbf{Diverse Semifactual Explanations of Reject (DSER)} Artelt \& Hammer \cite{artelt2022even} proposed a semi-factual method based on optimization methods to explain reject decisions in machine learning; that is, to explain why a model should \textit{not} make a prediction. It applies its loss function using four constraints for good semi-factuals (i.e., feasibility, sparsity, similarity, diversity):

\begin{equation}
\textit{DSER}(q)=\arg \min_{q_{sf} \epsilon \mathbb{R}^d} \ell(q_{sf})
\end{equation}

\noindent where, $q_{sf}$ is the semi-factual of query $q$ and $\ell()$ represents the combined loss function such that, 

\begin{equation}
    \ell(q_{sf})=\ell_{feasibile}(q_{sf})+\ell_{sparse}(q_{sf})+\ell_{similar}(q_{sf})+\ell_{diverse}(q_{sf})
\end{equation}

\noindent where feasibility is cast as,

\begin{equation}
\ell_{\text {feasible }}\left(q_{sf}\right) =C_{\text {feasible}} \cdot \max \left(r\left(q_{sf}\right)-\theta, 0\right)+C_{sf} \cdot \max \left(r(q)-r\left(q_{sf}\right), 0\right)
\end{equation}

\noindent which ensures that the semi-factual is also predictively uncertain but more certain than the original query, \emph{q} (to be convincing). Here, $C$ represents the regularization parameters for each component, $r()$ is the reject function based on the certainty of predictive function and $\theta$ is the reject threshold. 

\begin{equation}
\ell_{\text {sparse}}\left(q_{sf}\right)=C_{\text {sparse}} \cdot \max \left(\sum_{i=1}^d \mathbbm{1}\left(\left(q_{sf}-q\right)_i \neq 0\right)-\mu, 0\right)
\end{equation}

\noindent covers \textit{sparsity} promoting candidates with fewer feature differences between the semi-factual and the query. Here $d$ is the number of feature dimensions and $\mu \geq 1$ is a hyperparameter that controls the number of feature-differences.  

\begin{equation}
\ell_{\text {similar}}\left(q_{sf}\right)=-C_{\text {similar }} \cdot\left\|q_{sf}-q\right\|_2
\end{equation}

\noindent deals with \textit{similarity} promoting greater distance between the query and the semi-factual in Euclidean space, and finally,

\begin{equation}
\ell_{\text {diverse }}\left(q_{sf}\right)=C_{\text {diverse }} \cdot \sum_{j \in \mathcal{F}} \mathbbm{1}\left(\left(q_{sf}-q\right) \neq 0\right)
\end{equation}

\noindent handles \textit{diversity} ensuring that several featurally-distinct semi-factuals are generated. Here, $\mathcal{F}$ represents the set of features that have already been used to generate semi-factuals, feature-sets that should be avoided:

\begin{equation}
\mathcal{F}=\left\{j \mid \exists i:\left(q_{sf}^i-q\right)_j \neq 0\right\}
\end{equation}

    \item \textbf{PlausIble Exceptionality-based Contrastive Explanations (PIECE)} This method proposed by Kenny \& Keane \cite{kenny2021generating} computes semi-factuals ``on the way to'' computing counterfactuals using statistical techniques and a Generative Adverserial Network (GAN) model. It identifies "exceptional" (i.e., probabilistically-low) features in a query with respect to its counterfactual class and iteratively modifies these features until they become "normal" (i.e., probabilistically-high). As these exceptional features are incrementally altered, the generated instances gradually move away from the query towards the counterfactual class, with the last instance just before the decision boundary being selected to be the semi-factual. So, the semi-factual is like a point on the trajectory from the query to the counterfactual.

    Exceptional features are identified using the statistical probabilities in the training distribution of the counterfactual class $c^{\prime}$. Specifically, a two-part hurdle process is used to model the latent features of the query (when it is an image) in the feature-extracted layer ($X$) of a Convolutional Neural Network (CNN) with an ReLU activation function. The first hurdle process is modelled as a Bernoulli distribution and the second as a probability density function (PDF) as: 

\begin{equation}
    p\left(x_i\right)=\left(1-\theta_i\right) \delta_{\left(x_i\right)(0)}+\theta_i f_i\left(x_i\right), \quad \text { s.t. } \quad x_i \geq 0
\end{equation}

\noindent where $p(x_i)$ is the probability of the latent feature value $x_i$ for $c^{\prime}$, $\theta_i$ is the probability of the neuron in $X$ activating for $c^{\prime}$ (initial hurdle process), and $f_i$ is the subsequent PDF modelled (the second hurdle process). The constraint of $x_i \geq 0$ refers to the ReLU activations, and $\delta_{\left(x_i\right)(0)}$ is the Kronecker delta function, returning 0 for $x_i > 0$, and 1 for $x_i = 0$.  After modelling the distribution, a feature value $x_i$ is regarded as an exceptional feature for the query  in situations where, 

\begin{equation}
x_i=0 \mid p\left(1-\theta_i\right)<\alpha
\end{equation}

\noindent if the neuron $X_i$ does not activate, given the probability of it not activating being less than $\alpha$ for $c^{\prime}$, and,

\begin{equation}
x_i>0 \mid p\left(\theta_i\right)<\alpha
\end{equation}

\noindent if a neuron activates, given that the probability of it activating being less than $\alpha$ for $c^{\prime}$, where $\alpha$ is a threshold.

Once the exceptional features are identified, the query's features are adjusted to their expected values ($x^{\prime}$) with generated instances being checked by the CNN to be in the query or counterfactual-class. The semi-factual is the last generated instance in the query-class before crossing into the counterfactual-class. Finally, a GAN is used to visualize the explanations by identifying a latent vector ($z^{\prime}$) such that loss between $x^{\prime}$ and $C(G(z^{\prime}))$ is minimized as,

\begin{equation}
z^{\prime}=\underset{z}{\arg \min }\left\|C(G(z))-x^{\prime}\right\|_2^2
\end{equation}

\begin{equation}
    \textit{PIECE}(q) = G(z^{\prime})
\end{equation}

\noindent where $C$ is a CNN classifier and $G$ is the GAN generator.

    \item \textbf{Class-to-Class Variational Autoencoder (C2C-VAE)} Zhao et. al \cite{zhao2022generating} proposed an efficient way to obtain semi-factuals based on autoencoders. They used a variational autoencoder (VAE) with an encoder $(f)$ and a decoder $(f^{\prime})$ to learn the embedding space representing the differences between feature patterns in two classes. In the initial learning phase, given a pair of cases $s$ and $t$ from two classes, C2C-VAE encodes the feature difference, $f_{\Delta}$, where $f_{\Delta}(s, t)=f(s)-f(t)$ using an encoder $g$, as $g\left(<f_{\Delta}, C_s, C_t>\right)$ and decodes the embedding using a decoder $g^{\prime}$ as $f_{\Delta}^{\prime}=g^{\prime}\left(g\left(<f_{\Delta}, C_s, C_t>\right)\right.$

    To derive an explanation for a query, $q$ in class $C_q$, the method first generates a guide $t$ in the counterfactual class $C_t$. This guide selection leverages the feature difference embedding space $g$. Specifically, the method randomly samples vectors from $g$, decodes them back to the original feature space, and selects the one with the least mean squared error compared to $q$. Finally, it interpolates between the extracted features of $f(q)$ and $f(t)$ in the VAE's latent space to obtain semi-factuals as, 

\begin{equation}
C2C\text-VAE(q)=f^{\prime}((1-\lambda) * f(q)+\lambda * f(t)),  0 \leq \lambda \leq 1
\end{equation}

\noindent where $\lambda$ is a hyperparameter which determines the weight of interpolation between $q$ and $t$ and controls whether the output is more similar to $q$ (for a semi-factual) or $t$ (for a counterfactual). 

    \item \textbf{Most Distant Neighbor (MDN)} Aryal \& Keane \cite{aryal2024even} proposed and this method as a na\"ive, benchmark algorithm to find query-class instances that are most distant from the query on some dimension(s), while also sharing many common features. MDN scores all the query-class's instances on the extremity of their feature-values, determining whether they are much higher or lower than the feature-values of the query, \textit{q}, to find its most distant neighbor. Its custom distance function, \textit{Semi-Factual Scoring} (\textit{sfs}), prioritises instances that are sparse relative to the query (i.e., fewer feature differences), but have the highest value-differences in their non-matching features, as follows:

\begin{equation}\label{eq:sfs}
\textit{sfs}(q, S, F)=\frac{same(q, x)}{F}+\frac{\textit{diff}(q_f, x_f)} {\textit{diff}_{max}(q_f, S_f)}
\end{equation}

\noindent where \textit{S} is Higher/Lower Set and $x \in S$, \textit{same()} counts the number of features that are equal between \textit{q} and \textit{x}, \textit{F} is the total number of features, \textit{diff()} gives the difference-value of key-feature, \textit{f}, and \textit{diff}$_{max}()$ is the maximum difference-value for that key-feature in the Higher/Lower Set. The best-feature-MDN is selected as the instance with the highest \textit{sfs} score from the Higher/Lower set for each feature, independently. Finally, the best of the best-feature-MDNs across all dimensions is chosen as the overall semi-factual for the query,

\begin{equation}
\textit{MDN}(q, S)=\arg \max_{x \in S} sfs(x)
\end{equation}

    \item \textbf{Explanations for Positive Outcomes (S-GEN)} Kenny \& Huang \cite{kenny2023sf} proposed the novel concept of ``gain" (akin to ``cost" in counterfactuals) as a new constraint for semi-factual methods. They argue that semi-factuals best explain positive outcomes, whereas counterfactuals work best for negative outcomes. Their S-GEN method uses \textit{gain}, along with traditional constraints (such as plausibility, robustness and diversity) to compute semi-factuals that inform users about better and positive recourses. It's objective function is: 

\begin{equation}\label{eq:sfpositive}
\begin{aligned}
S\text-GEN(\mathbf{q})=\max _{\mathbf{a}_1, \ldots, \mathbf{a}_m} & \frac{1}{m} \sum_{i=1}^m f\left(P\left(\mathbf{q}, \mathbf{a}_i\right), G\left(\mathbf{q}, \mathbf{a}_i\right)\right)+\gamma R\left(\left\{\mathbf{q}_1^{\prime}, \ldots, \mathbf{q}_m^{\prime}\right\}\right) \\
\text { s.t. } & \forall i, j: \mathbf{q}_i^{\prime}=S_{\mathcal{M}}\left(\mathbf{q}, \mathbf{a}_i\right), H_j\left(\mathbf{q}_i^{\prime}\right) \geq 0(\text { or }>)
\end{aligned}
\end{equation}

\noindent where, $\mathbf{a_i}$ represents an action taken on $i^{th}$ feature-dimension, $m$ is the desired number of semi-factuals to be generated. $
P(\mathbf{q}, \mathbf{a}_i)=\operatorname{Pr}\left(S_{\mathcal{M}}(\mathbf{q}, \mathbf{a}_i)\right)
$ denotes the plausibility of explanation for $\mathbf{q}$ by taking action $\mathbf{a}_i$ where $Pr$ is the distribution density. S-GEN uses a Structural Causal Model (SCM), $S_{\mathcal{M}}$ to better capture the causal dependencies between the features and hence obtain feasible explanations.

\begin{equation}
G(\mathbf{q}, \mathbf{a}_i)=\mathcal{P}_{SF} \circ \delta\left(\mathbf{q}, S_{\mathcal{M}}(\mathbf{q}, \mathbf{a}_i)\right)
\end{equation}

\noindent represents the gain function for $\mathbf{q}$ by taking action $\mathbf{a}_i$ where $\delta()$ is the distance function. Intuitively, it measures the difference between original state $\mathbf{q}$ and the new state obtained by the transition from $\mathbf{q}$ by taking action $\mathbf{a}_i$ through an SCM, $S_{\mathcal{M}}(\mathbf{q},\mathbf{a}_i)$. Greater differences indicate higher ``gains'' (i.e., better explanations).

\begin{equation}
R\left(\left\{\mathbf{q}_i\right\}_{i=1}^m\right)=\frac{2}{m(m-1)} \sum_{i=1}^m \sum_{j>i}^m L_p \circ \delta\left(\mathbf{q}_i, \mathbf{q}_j\right)
\end{equation}

\noindent shows the diversity function which is regularized by $\gamma$ in Eq. (\ref{eq:sfpositive}).  $L_p$ is the \mbox{$L_p$-norm} and $\delta()$ is the distance function used, so as many distinct explanations as possible are generated that are also far from each other. Finally,  robustness is achieved in a post-hoc manner as a hard constraint defined by:

\begin{equation}
H(\mathbf{q}, \mathbf{a})=\min _{\mathbf{q}^{\prime} \in \mathbb{B}_s(\mathbf{q}, \mathbf{a})} h\left(\mathbf{q}^{\prime}\right)-\psi
\end{equation}

\noindent where $H(\mathbf{q}, \mathbf{a})$ denotes the post-robustness of an action $\mathbf{a}$ for a test instance $\mathbf{q}$. The intuition is that any instances lying in the neighborhood $\mathbb{B}_s$ of the generated semi-factual $\mathbf{q}^\prime=S_{\mathcal{M}}(\mathbf{q},\mathbf{a})$ after taking the action $\mathbf{a}$ also have a positive outcome. This function ensures that the output of a predictive model $h$ for $\mathbf{q}^\prime$ is higher than a threshold $\psi=0.5$ (in case of binary classes).
    
    \item \textbf{Diverse Counterfactual Explanations (DiCE)} Mothilal et. al \cite{mothilal2020explaining} used an optimization method based on various constraints such as distance, diversity (using determinantal point processes) and feasibility (using causal information from users) to compute counterfactuals which can be used to obtain semi-factuals by adjustments to the loss function: 

\begin{equation}
\begin{split}
DiCE(q)&=\underset{\boldsymbol{c}_1, \ldots, c_k}{\arg \min } \frac{1}{k} \sum_{i=1}^k \operatorname{yloss}\left(f\left(c_i\right), y\right)+\frac{\lambda_1}{k} \sum_{i=1}^k \operatorname{dist}\left(c_i, q\right) \\
&\qquad {}- \lambda_2 \text {dpp\_diversity}(c_1,...,c_k)
\end{split}
\end{equation}

\noindent where $q$ is the query input, $c_i$ is a counterfactual explanation, $k$ is the total number of diverse counterfactuals to be generated, $f()$ is the black box ML model, $yloss()$ is the metric that minimizes the distance between $f()$'s  prediction for $c_i$ and the desired outcome $y$, $dist()$ is the distance measure between $c_i$ and $q$, and dpp\_diversity() is the diversity metric. $\lambda_1$ and $\lambda_2$ are hyperparameters that balance the three components of the loss function.
    
\end{itemize}

\textbf{Metrics.} The following 5 key metrics were used to assess the quality of semi-factuals generated from each method. 

\begin{itemize}[itemsep=2ex]
    \item \textit{Distance.} It measures the $L_2$-norm distance between query and the semi-factual where higher is better. 
    
    \item \textit{Sparsity.} It measures the number of feature difference between query and the semi-factual as a ratio of desired (set to 1) to observed-difference where higher value is preferred. 
    \begin{equation}
Sparsity  = \frac{desired_{\textit{diff}}}{observed_{\textit{diff}}}
\end{equation}

    \item \textit{Plausibility.} It measures the distance between the semi-factual and the nearest training instance where lower value is better. 
    
    \item \textit{Trustworthiness.} It measures the confidence for the semi-factual being in the query class compared to the counterfactual class measured as a ratio of their class distance where higher score is better. 
    \begin{equation}\label{eq:confusability}
\textit{Trustworthiness(x)}=\frac{\textit{d(x,CF)}}{\textit{d(x,Q)}}
\end{equation}
\noindent where $x$ is the semi-factual, \textit{CF} is the counterfactual-class, $Q$ is the query-class and $d()$ measures the distance.

    \item \textit{Robustness.} It measures the difference in semi-factuals obtained through small perturbations of the query measured using Lipschitz continuity where lower value is better.
    \begin{equation}\label{eq:robustness}
Robustness(x)=\underset{x_i \in B_\epsilon\left(x\right)}{\operatorname{argmax}} \frac{\left\|f\left(x\right)-f\left(x_i\right)\right\|_2}{\left\|x-x_i\right\|_2}
\end{equation}

\noindent where $x$ is the input query, $B_\epsilon\left(x\right)$ is the ball of radius $\epsilon$ centered at $x$, $x_i$ is a perturbed instance of $x$ and $f()$ is the explanation method. 

\end{itemize}

\textbf{Datasets. } The evaluation was carried out across 7 benchmark tabular datasets which are all binary-classed. 
\begin{itemize}
    \item Adult Income (N=26,540, 12 features)
    \item Blood Alcohol (N=2000, 5 features)
    \item PIMA Diabetes (N=392, 8 features)
    \item Default Credit Card (N=30,000, 23 features)
    \item German Credit (N=1000, 20 features)
    \item HELOC (N=8291, 20 features)
    \item Lending Club (N=39,239, 8 features)
\end{itemize}

\textbf{Procedure.} For a given query in a dataset, its corresponding semi-factual was computed using all the 8 semi-factual methods along with their respective values for each evaluation metric. The semi-factual which had the highest aggregated score combining all the metrics was determined to be the best semi-factual explanation for the query. This process was carried out across all datasets to obtain best semi-factuals for each query-instance. A sample of best semi-factuals from each dataset was then used to analyze the seesaw pattern.

\textbf{Results.} The results in Table \ref{tab1} show that out of 10,300 best semi-factuals sampled, 9175 (or 89\%) of them revealed the seesaw pattern of changing marginal contributions between key- and hidden-features. This validates our intuition that this property exists in most good semi-factuals. 

\begin{table}[h]
\centering
\caption{Table showing the number of best semi-factuals sampled from each dataset and how many of those had the seesaw pattern.}\label{tab1}
\begin{tabular}{llll}
\toprule
Dataset &  \makecell[l]{Sample of \\Best Semi-factuals} & \makecell[l]{Best-Semi-factuals \\with Seesaw Pattern} \\
\midrule
Adult Income & 1200 & 1029 \\
Blood Alcohol & 400 & 360 \\
PIMA Diabetes & 100 & 92 \\
Default Credit Card & 3200 & 2826 \\
German Credit & 400 & 358 \\
HELOC & 2000 & 1820 \\
Lending Club & 3000 & 2690 \\
\midrule
Total & 10,300 & 9,175 \\
\bottomrule
\end{tabular}
\end{table}

\subsection{Baseline Implementation}
In this section, we discuss the implementation details and parameter specifications for the existing semi-factual methods. 

\begin{itemize}[itemsep=2ex]
\item \textbf{Local Region Model} The local surrogate model was trained with a minimum of 200 instances from each class. 
\item \textbf{KLEOR} A k-NN model with k=3 was used to compute the distances between the instances and obtain NUN.
\item \textbf{DSER} The model was implemented based on the publicly available library\footnote{\url{https://github.com/HammerLabML/DiverseSemifactualsReject}}. A k-NN classifier was used to fit the conformal predictor with k=5. The reject threshold $\theta$ was set to 0.4. The regularization parameter $C$ for each component in the loss function was set to 1. The hyperparamter $\mu$ to control the number of feature-differences was set to 2.
\item \textbf{PIECE} The framework was modified to work with the tabular data inspired by Kenny \& Huang's \cite{kenny2023sf} modification. The training data was partitioned into two subsets based on model predictions: instances predicted as the original class $c$ and those predicted as the counterfactual class $c'$. For each subset, feature distributions were modeled independently, using Beta distributions for continuous features and categorical distributions for discrete features. To construct a semi-factual predicted as $c$, we evaluate the probability of each query feature value under the distributions of class $c'$. Features are then sequentially adjusted to their expected values under $c'$, starting from the lowest-probability feature, until the next modification would cross the decision boundary. For continuous features, the probability of a value is defined as the minimum of the two integrals on either side of that value in the distribution. If an expected value lies outside the permitted actionability range, it is clipped to the nearest feasible value.
\item \textbf{C2C-VAE} To adapt to the tabular data, the encoder and decoder in both VAE and C2C-VAE were implemented using 3 and 2 fully connected layers, respectively, with the latent dimension $z$ set to 4. The value of interpolation parameter $\lambda$ was set to 0.2. 
\item \textbf{MDN} In the \textit{sfs()} function, the categorical features were determined to be similar by direct comparison of their values whereas the continuous features were considered to be the same if the values lie within $\pm20\%$ of standard deviation of the selected feature.  
\item \textbf{S-GEN} We followed the publicly available implementation of S-GEN\footnote{\url{https://github.com/EoinKenny/Semifactual_Recourse_Generation}} with default actionability constraints and hyperparameter specifications for causal and non-causal settings.
\item \textbf{DiCE} The DiCE method was implemented using the publicly available library\footnote{\url{https://github.com/interpretml/DiCE}}. The random search method was used and all parameters were kept at their default settings except the $desired\_class$ was set to "same" in the objective function to ensure the generation of semi-factuals. 
\end{itemize}

\subsection{NSGA-II Implementation in ISF}
The NSGA-II algorithm used to solve the constrained multi-objective optimization in ISF was implemented via the pymoo\footnote{\url{https://github.com/anyoptimization/pymoo}} framework. The algorithm was initialized using random float sampling and evolved using standard evolutionary operators. Parent selection was performed using tournament selection, while offspring were generated using simulated binary crossover (SBX) (crossover probability = 0.9, distribution index = 15) and polynomial mutation (mutation probability = 0.9, distribution index = 20). Environmental selection followed the standard NSGA-II rank-and-crowding survival strategy, which performs non-dominated sorting and maintains diversity using crowding distance. Duplicate solutions were eliminated during evolution. The algorithm was executed with a population size of 50 for 100 generations, while all other parameters were kept at their default values provided in the pymoo implementation.

\subsection{User Studies}
The materials for the user studies were based on German Credit dataset. We used 5 important features and modified the dependent variable (class) to be loan accepted and rejected rather than the original good and bad credit. The semi-factual explanations in both user studies were generated using ISF method. Below we show the initial setup of both user studies before showing some sample scenarios for each study. 

\begin{center}
    \includegraphics[width=1.\linewidth]{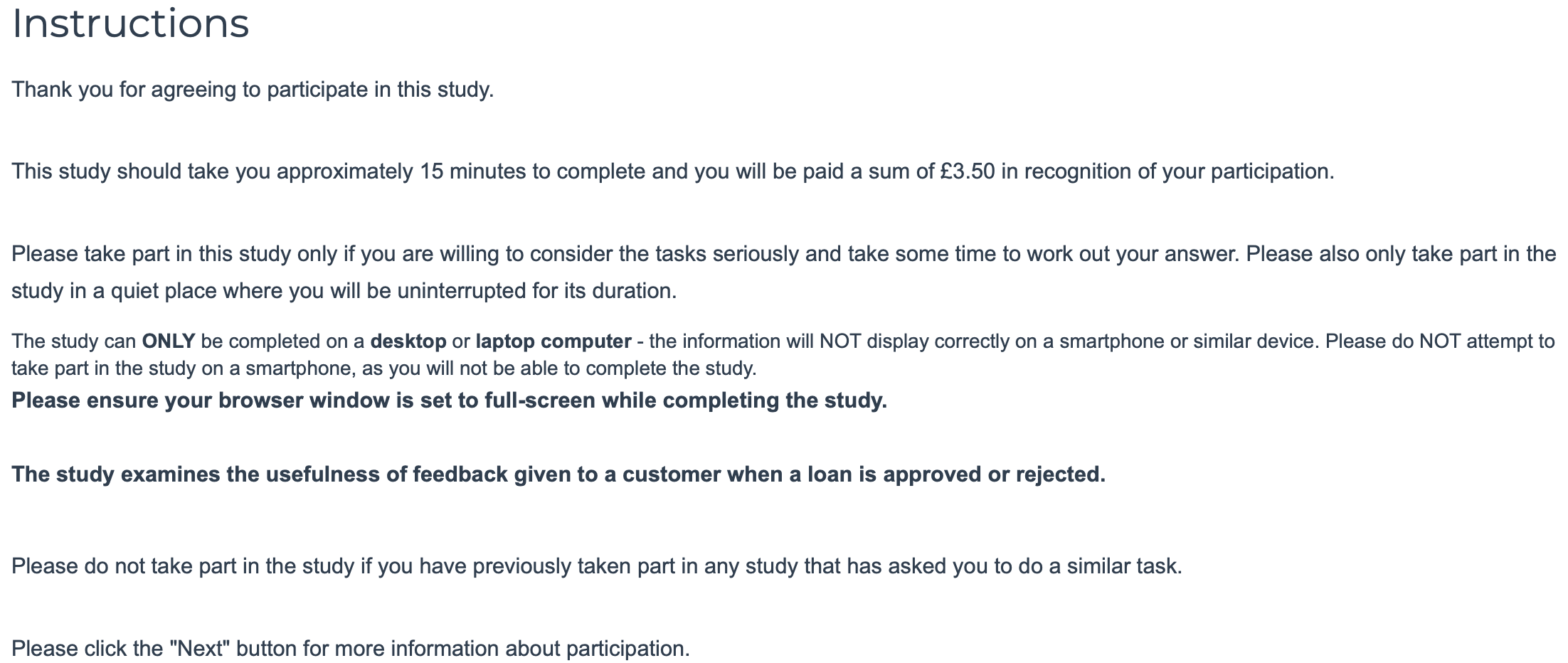}
    \includegraphics[width=1.\linewidth]{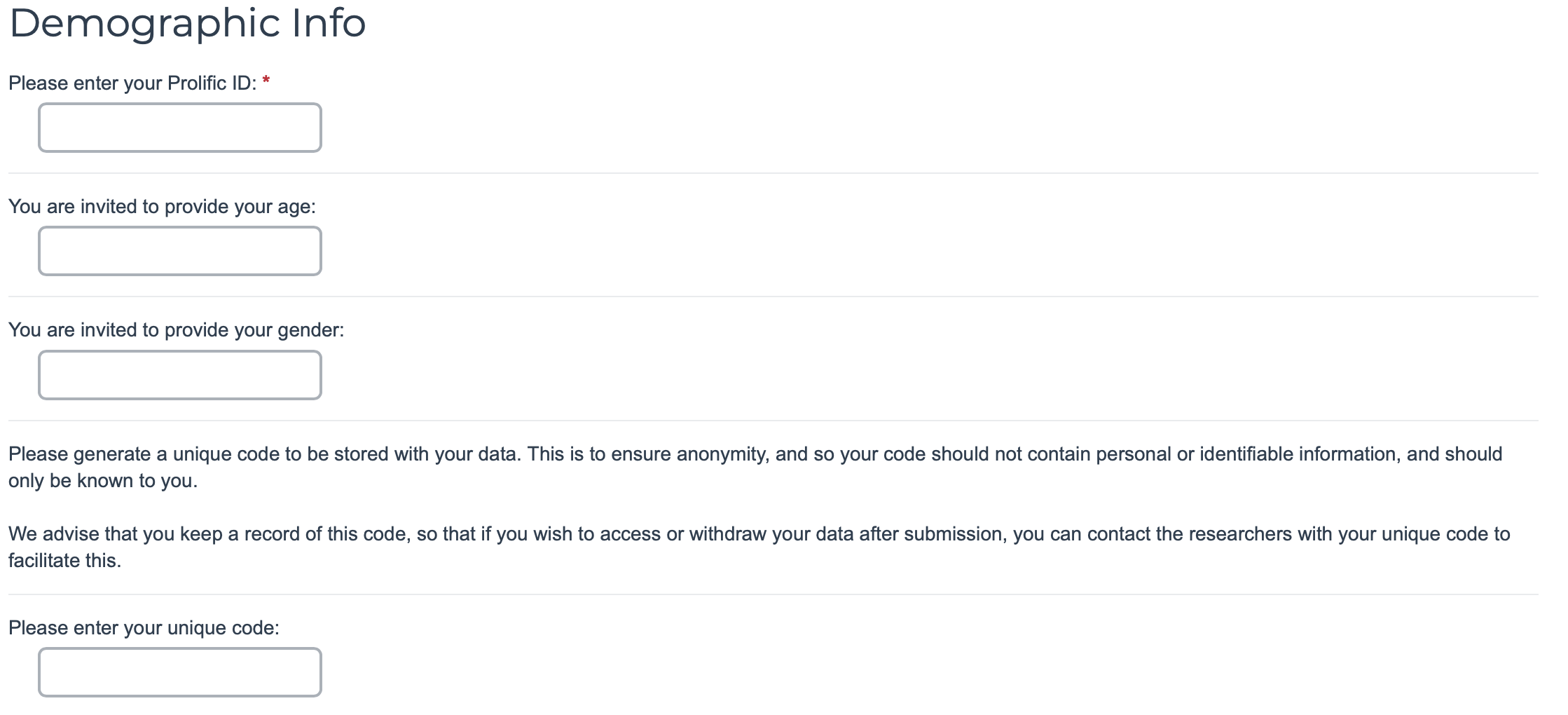}
    \includegraphics[width=1.\linewidth]{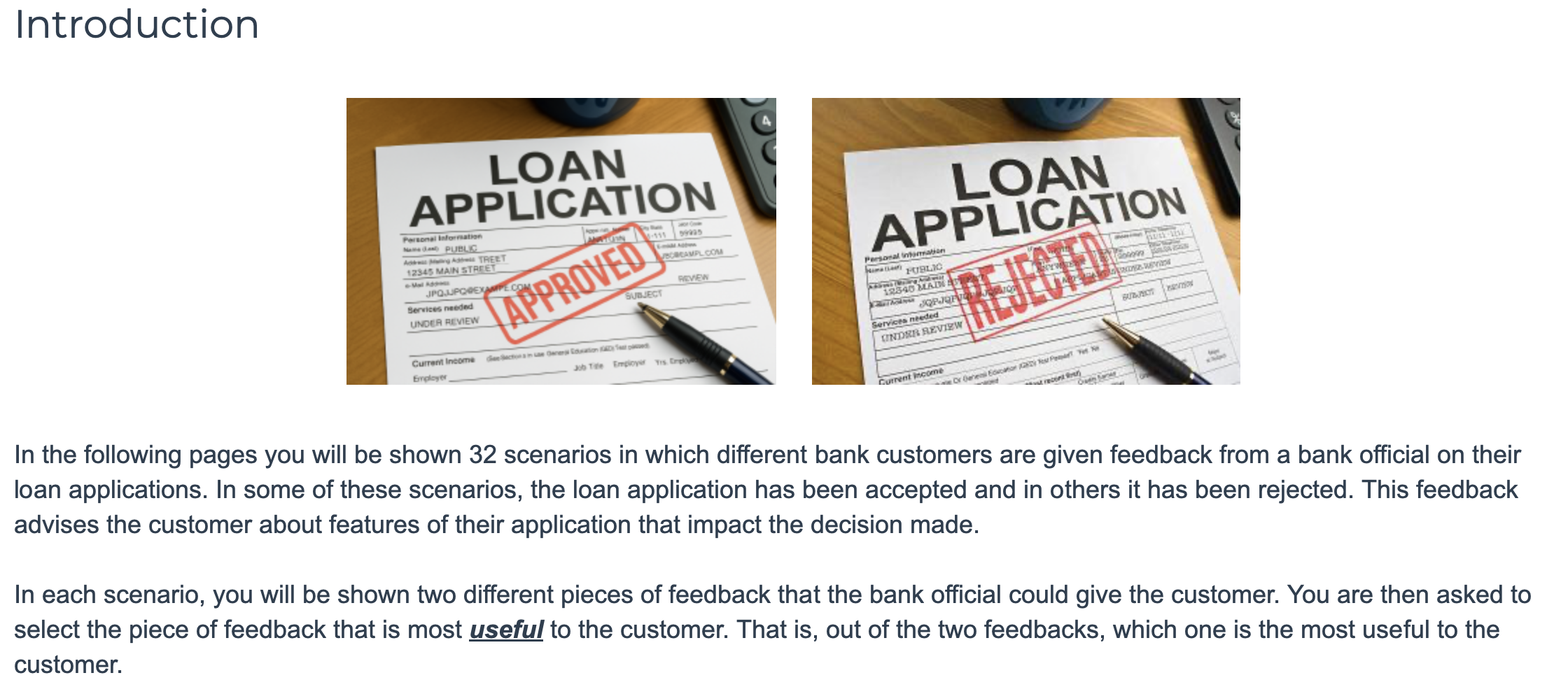}
    \includegraphics[width=1.\linewidth]{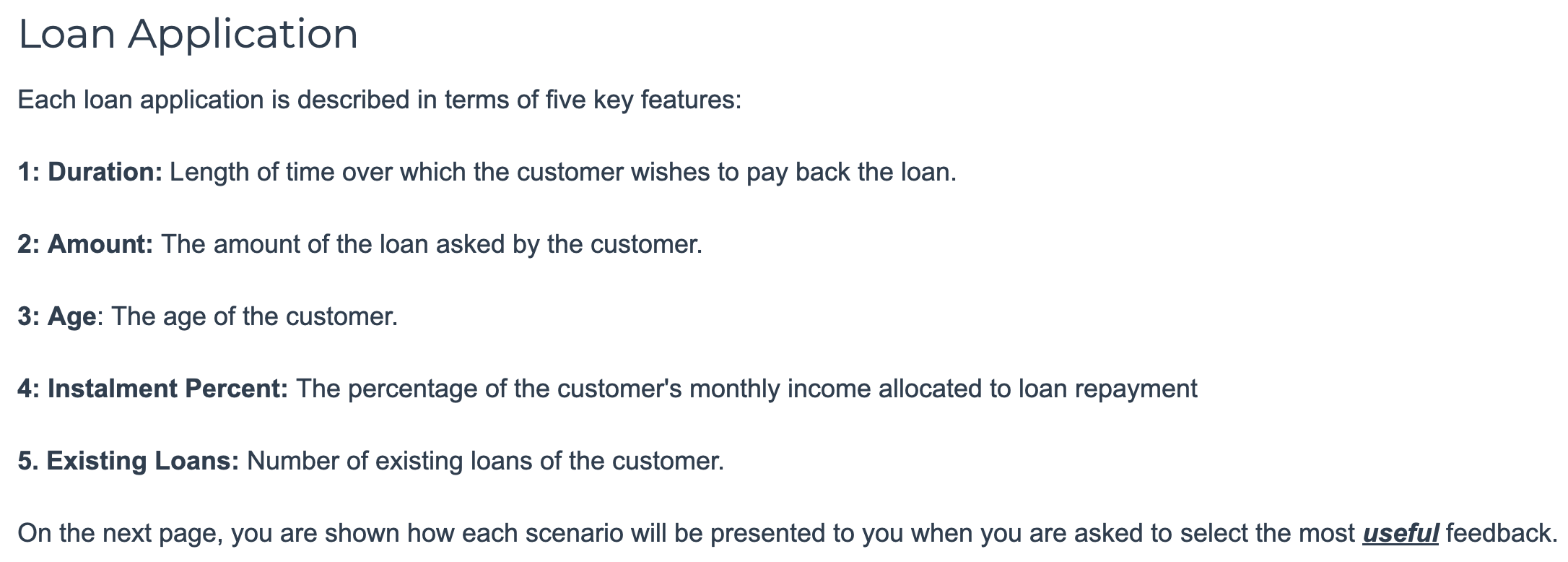}
    \includegraphics[width=1.\linewidth]{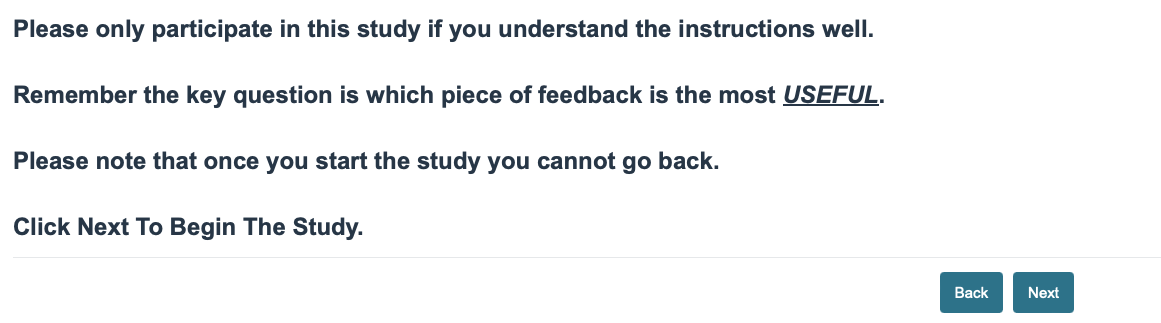}
\end{center}

\clearpage
\subsubsection{User Study 1: Do People Prefer Informative Explanations?}

\begin{center}
    Sample Scenario 1 \includegraphics[width=1.\linewidth]{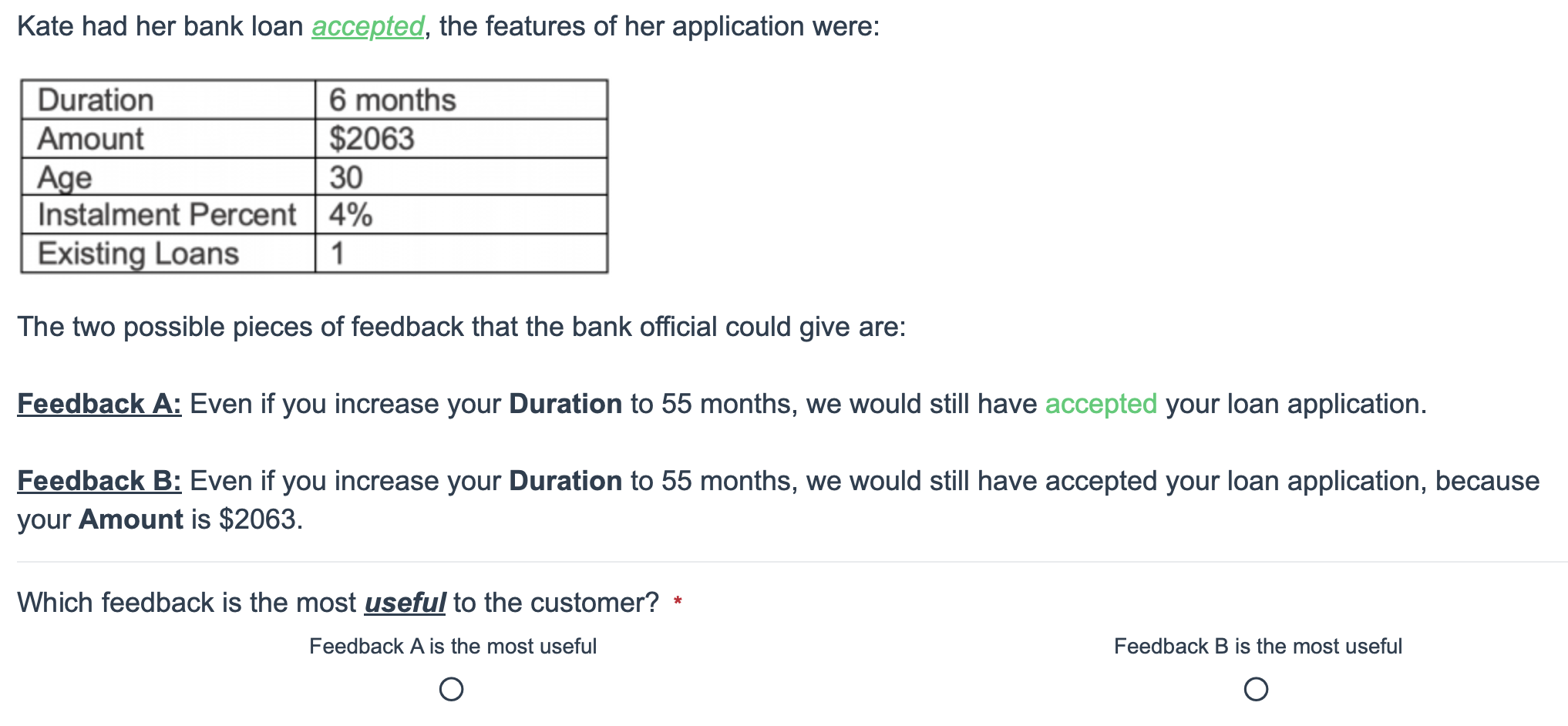}
    Sample Scenario 2 \includegraphics[width=1.\linewidth]{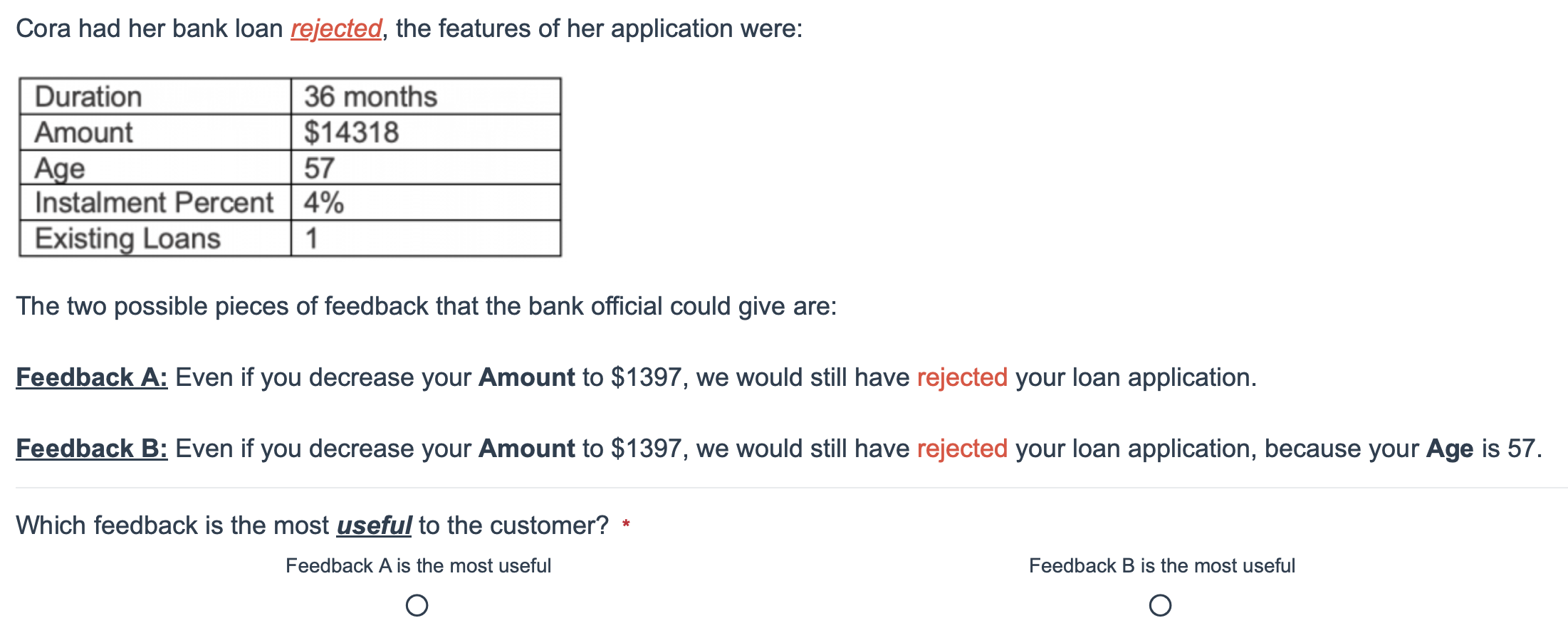}
\end{center}

\pagebreak
\subsubsection{User Study 2: Preferring Good or Bad Explanations?} 

\begin{center}
    Sample Scenario 1 \includegraphics[width=1.\linewidth]{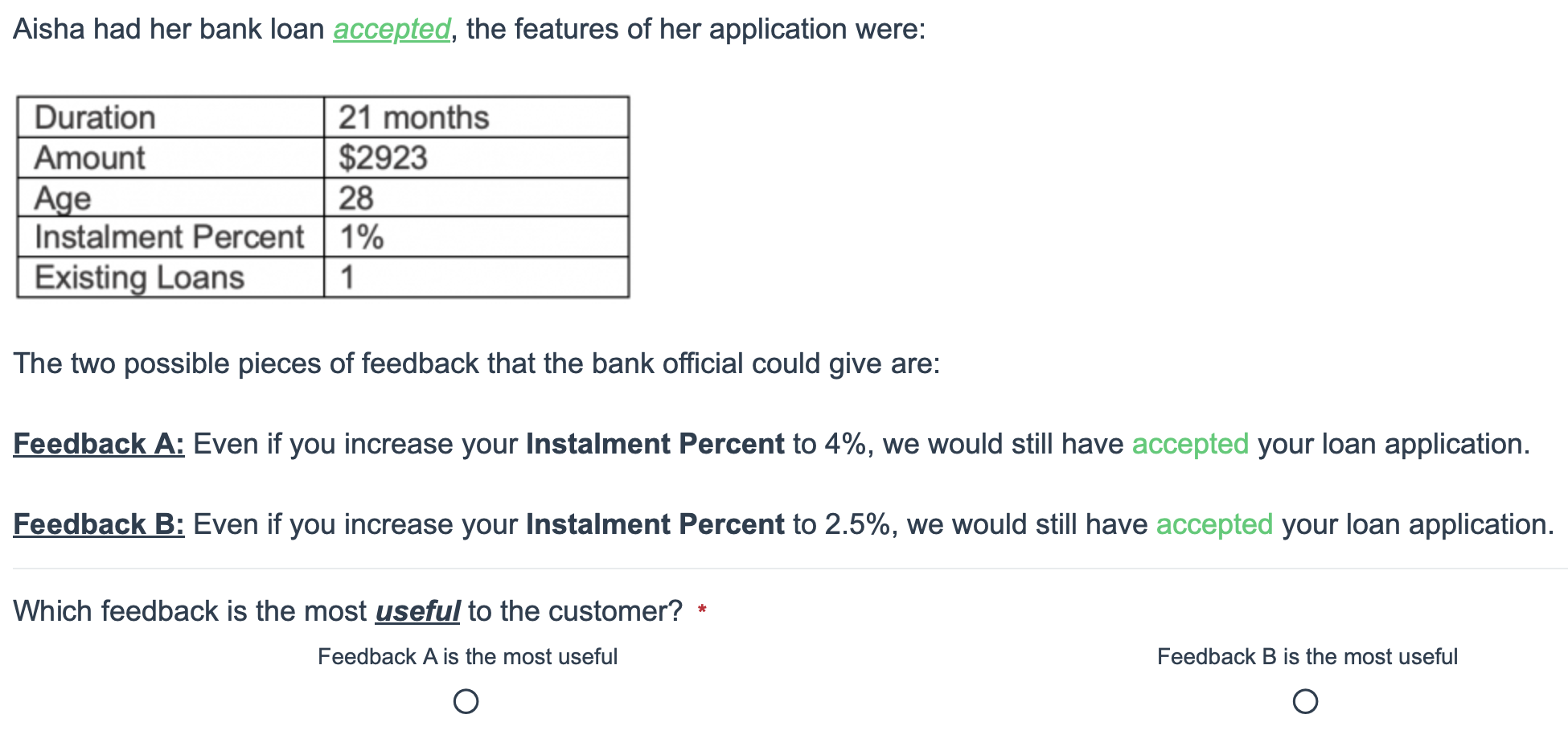}
    Sample Scenario 2 \includegraphics[width=1.\linewidth]{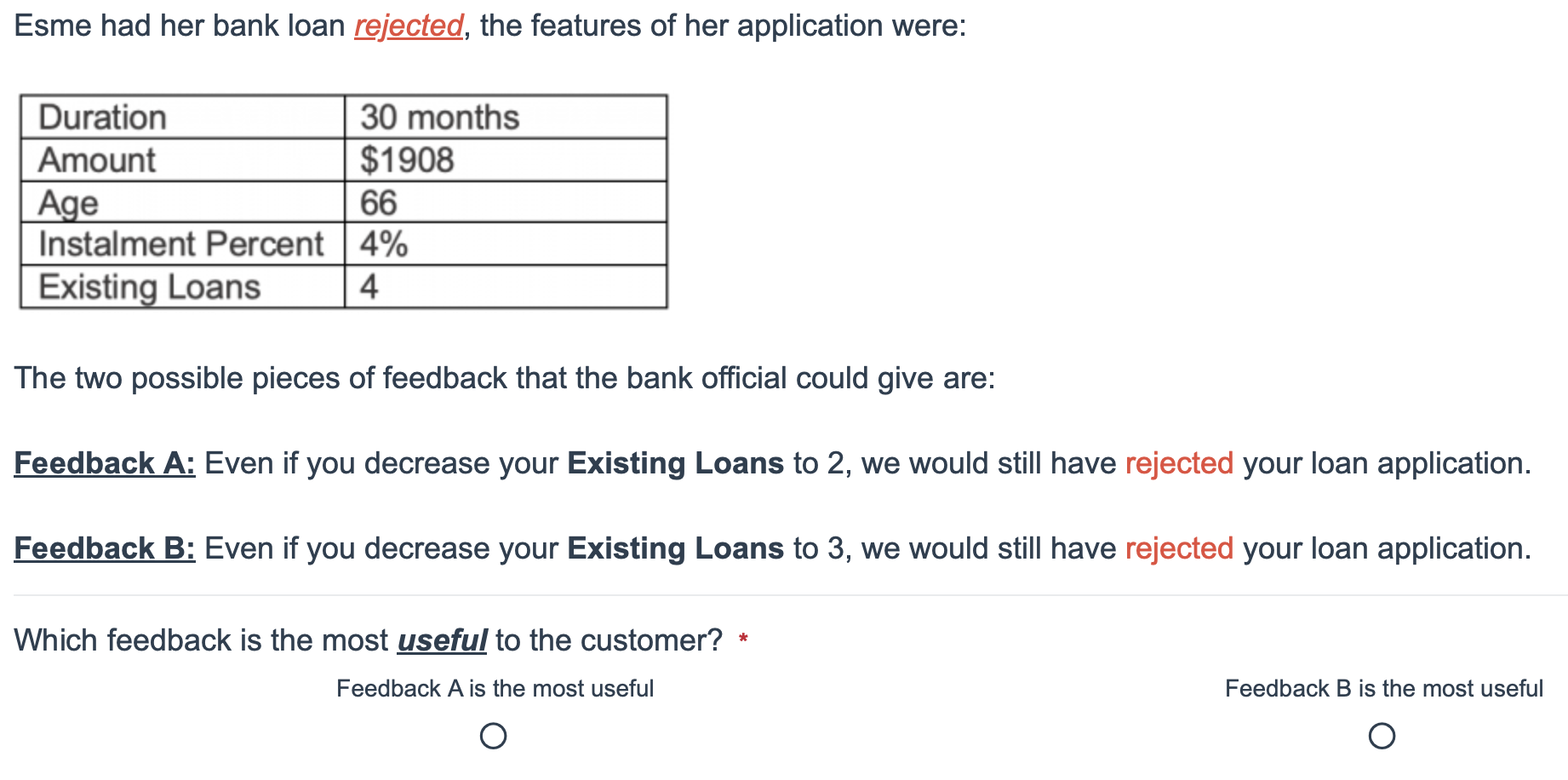}
\end{center}

\end{appendix}

\end{document}